\documentclass[10pt,twocolumn,letterpaper]{article}

\usepackage{cvpr}              

\usepackage{graphicx}
\usepackage{amsmath}
\usepackage{amssymb}
\usepackage{booktabs}
 \usepackage{multirow}
\usepackage{enumitem}
\usepackage[title]{appendix}

\usepackage[table,xcdraw]{xcolor}

\usepackage[pagebackref,breaklinks,colorlinks]{hyperref}

\usepackage[capitalize]{cleveref}
\crefname{section}{Sec.}{Secs.}
\Crefname{section}{Section}{Sections}
\Crefname{table}{Table}{Tables}
\crefname{table}{Tab.}{Tabs.}

\def\cvprPaperID{4855} 
\def\confName{CVPR}
\def\confYear{2023}

\makeatletter
\def\thanks#1{\protected@xdef\@thanks{\@thanks
        \protect\footnotetext{#1}}}
\makeatother

\begin{document}

\title{Visual-Language Prompt Tuning with Knowledge-guided Context Optimization}

\author{\small Hantao Yao$^1$,Rui Zhang$^2$, Changsheng Xu$^{1,3}$\\
$^1$ \small State Key Laboratory of Multimodal Artificial Intelligence Systems, Institute of Automation, CAS\\
$^2$ \small State Key Lab of Processors, Institute of Computing Technology, CAS; $^3$ \small University of Chinese Academy of Sciences(CAS),\\
{\tt\small hantao.yao@nlpr.ia.ac.cn}
}

\maketitle

\begin{abstract}
Prompt tuning is an effective way to adapt the pretrained visual-language model (VLM) to the downstream task using task-related textual tokens.
Representative CoOp-based work combines the learnable textual tokens with the class tokens to obtain specific textual knowledge.
However, the specific textual knowledge is worse generalization to the unseen classes because it forgets the essential general textual knowledge having a strong generalization ability.
To tackle this issue, we introduce a novel Knowledge-guided Context Optimization (KgCoOp) to enhance the generalization ability of the learnable prompt for unseen classes. 
The key insight of KgCoOp is that the forgetting about essential knowledge can be alleviated by reducing the discrepancy between the learnable prompt and the hand-crafted prompt.
Especially, KgCoOp minimizes the discrepancy between the textual embeddings generated by learned prompts and the hand-crafted prompts.
Finally,  adding the KgCoOp upon the contrastive loss can make a discriminative prompt for both seen and unseen tasks.
Extensive evaluation of several benchmarks demonstrates that the proposed Knowledge-guided Context Optimization is an efficient method for prompt tuning, \emph{i.e.,} achieves better performance with less training time.
\href{https://github.com/htyao89/KgCoOp}{code}.
\end{abstract}

\section{Introduction}
\label{sec:intro}
With the help of the large scale of the image-text association pairs, the trained visual-language model (VLM) contains essential general knowledge, which has a better generalization ability for the other tasks.
Recently, many visual-language models have been proposed, such as Contrastive Language-Image Pretraining (CLIP)~\cite{RadfordKHRGASAM21}, Flamingo~\cite{abs-2204-14198}, ALIGN~\cite{JiaYXCPPLSLD21}, \emph{etc}. 
Although VLM is an effective model for extracting the visual and text description, training VLM needs a large scale of high-quality datasets.
However, collecting a large amount of data for training a task-related model in real visual-language tasks is difficult.
To address the above problem, the prompt tuning~\cite{ChoLTB21}~\cite{abs-2210-09263}~\cite{JiaTCCBHL22}~\cite{abs-2107-13586}~\cite{PetroniRRLBWM19}~\cite{RaoZ0TZH0L22}~\cite{TsimpoukelliMCE21}~\cite{abs-2109-11797} has been proposed to adapt the pretrained VLM to downstream tasks, achieving a fantastic performance on various few-shot or zero-shot visual recognization tasks.

\begin{table}[]
\caption{Compared to existing methods, the proposed KgCoOp is an efficient method, obtaining \textbf{a higher performance with less training time}.}
\label{tab:cem}
\centering
\footnotesize
\vspace{-0.8em}
\begin{tabular}{l|c|ccc|c}
\toprule
\multirow{2}{*}{Methods} & \multirow{2}{*}{Prompts} & \multicolumn{3}{c|}{Accuracy} &\multirow{2}{*}{Training-time} \\ \cline{3-5}
                         &                            & Base       & New       & H      &      \\
\midrule
\midrule
CLIP                     & hand-crafted &  69.34 & 74.22  &  71.70 & - \\
\midrule
CoOp                     & textual& \textbf{82.63}&67.99 & 74.60 &  6ms/image                   \\
ProGrad                  & textual&  82.48&   70.75& 76.16& 22ms/image                \\
CoCoOp                   &textual+visual& 80.47& 71.69 & 75.83 & 160ms/image              \\
\textbf{KgCoOp}                      &  textual &  80.73&   \textbf{73.6} & \textbf{77.0} &  \textbf{6ms}/image              \\
\bottomrule
\end{tabular}
\vspace{-1.5em}
\end{table}

The prompt tuning\footnote{In this work, we only consider the textual prompt tuning and do not involve the visual prompt tuning.} usually applies task-related textual tokens to embed task-specific textual knowledge for prediction.
The hand-crafted template ``a photo of a [Class]" in CLIP~\cite{RadfordKHRGASAM21} is used to model the textual-based class embedding for zero-shot prediction.
By defining the knowledge captured by the fixed (hand-crafted) prompts as the \emph{general textual knowledge}\footnote{Inspired from~\cite{HintonVD15}, `knowledge'in this work denotes the information contained in the trained model.}, it has a high generalization capability on unseen tasks.
However, the general textual knowledge is less able to describe the downstream tasks due to not consider the specific knowledge of each task.
To obtain discriminative task-specific knowledge, Context Optimization(CoOp)~\cite{ZhouYLL22}, Conditional Context Optimization(CoCoOp)~\cite{ZhouYL022}, and ProGrad~\cite{abs-2205-14865} replace the hand-crafted prompts with a set of learnable prompts inferred by the labeled few-shot samples.
Formally, the discriminative knowledge generated by the learned prompts is defined as the \emph{specific textual knowledge}.
However, CoOp-based methods have a worse generalization to the unseen classes with the same task, \emph{e.g.,} obtaining a worse performance than CLIP for the unseen classes (\emph{New}), shown in Table~\ref{tab:cem}.

As the specific textual knowledge is inferred from the labeled few-shot samples, it is discriminative for the seen classes and biased away from the unseen classes, leading to worse performance on the unseen domain.
For example, non-training CLIP obtains a higher \emph{New} accuracy on the unseen classes than CoOp-based methods, \emph{e.g.,} 74.22\%/63.22\%/71.69\% for CLIP/CoOP/CoCoOp.
The superior performance of CLIP on unseen classes verifies that its general textual knowledge has a better generalization for unseen classes.
However, the specific textual knowledge inferred by the CoOp-based methods always forgets the essential general textual knowledge, called catastrophic knowledge forgetting, \emph{i.e.,} the more serve catastrophic forgetting, the larger performance degradation.

\begin{figure}
  \centering
   \includegraphics[width=1.0\linewidth]{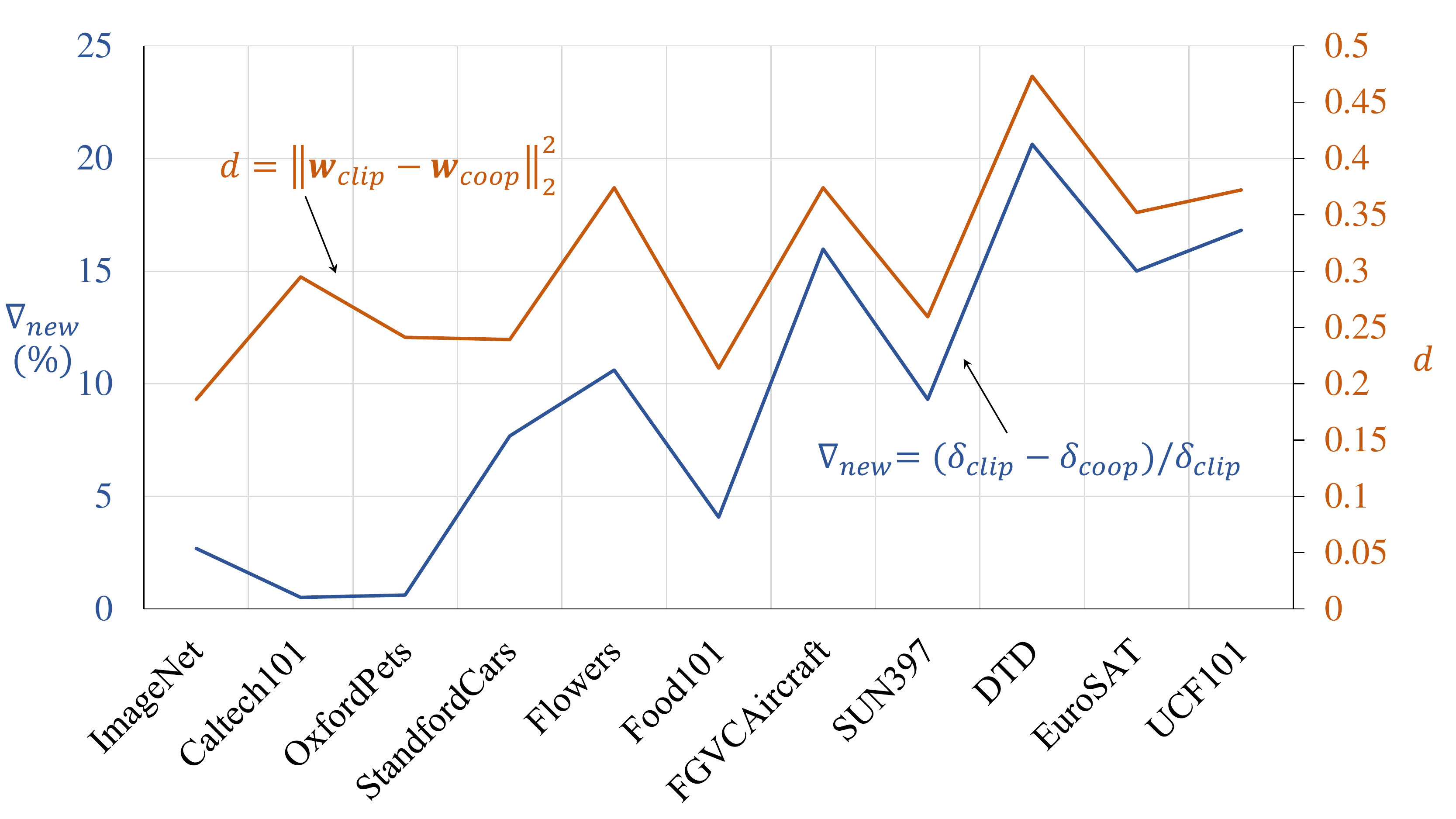}
   \caption{ For the CoOp-based prompt tuning, the degree of performance degradation $\triangledown_{new}$ on the \emph{New} classes is consistent with the distance between the learnable textual embedding $\mathbf{w}_{coop}$ and the  hand-crafted textual embedding $\mathbf{w}_{clip}$. The larger distance, the more severe the performance degradation. $\sigma_{clip}$ and $\sigma_{coop}$ are the accuracy of \emph{New} classes for CLIP and CoOp, respectively.}
   \label{fig:motivation}
\vspace{-1.0em}
\end{figure}

To address this issue, we introduce a novel prompt tuning method Knowledge-guided Context Optimization (KgCoOp) to boost the generality of the unseen class by reducing the forgetting of the general textual knowledge. The key insight of KgCoOp is that the forgetting about general textual knowledge can be alleviated by reducing the discrepancy between the learnable prompt and the handcrafted prompt.  
The observation of relationship between the discrepancy of two prompts and  the performance drop also verify the insight. As shown in  Figure~\ref{fig:motivation}, the larger the distance between textual embeddings generated by the learnable prompt and the hand-crafted prompt, the more severe the performance degradation.
Formally, the hand-crafted prompts ``a photo of a [Class]" are fed into the \emph{text encoder} of CLIP to generate the general textual embedding, regarded as the general textual knowledge.
Otherwise, a set of learnable prompts is optimized to generate task-specific textual embedding.
Furthermore, Knowledge-guided Context Optimization(KgCoOp) minimizes the euclidean distance between general textual embeddings and specific textual embeddings for remembering the essential general textual knowledge.
Similar to the CoOp and CoCoOp, the contrastive loss between the task-specific textual and visual embeddings is used to optimize the learnable prompts.

We conduct comprehensive experiments under base-to-new generalization setting, few-shot classification, and domain generalization over 11 image classification datasets and four types of ImageNets.
The evaluation shows that the proposed KgCoOp is an efficient method: \textbf{using the less training time obtains a higher performance,} shown in Table~\ref{tab:cem}.
In summary, the proposed KgCoOp obtains:
1) higher performance: KgCoOp obtains a higher final performance than existing methods. Especially, KgCoOp obtains a clear improvement on the \emph{New} class upon the CoOp, CoCoOp, and ProGrad, demonstrating the rationality and necessity of considering the general textual knowledge.
2) less training time: the training time of KgCoOp is the same as CoOp, which is faster than CoCoOp and ProGrad.


\section{Related Work}	
\paragraph{Visual-Language Models:} Recently, research has shown that using image-text association pairs can model a powerful visual-language model rather than merely considering the images.
The model inferred based on the image-text association pairs is defined as Visual-Language Model (VLM).
Recently, the visual-language models can be improved from the following aspects: 1) using a stronger text encoder or visual encoder, \emph{e.g.,} Transformers~\cite{VaswaniSPUJGKP17}; 2) contrastive representation learning~\cite{ChenK0H20}; 3) using more images~\cite{RadfordKHRGASAM21}~\cite{JiaYXCPPLSLD21}.
As training VLM needs a large-scale annotated dataset, unsupervised learning or weakly supervised learning~\cite{WangYYDT022} are used to train the visual-language model with the unannotated images.
Specially,  Masked Language Modeling (MLM) ~\cite{KimSK21}~\cite{LuBPL19} improves the robustness of visual and text embedding by randomly erasing the words in the text, and Masked autoencoders~\cite{HeCXLDG22}  is a scalable self-supervised learner by masking random patches of the input image.
As representive work is CLIP,  which trains the visual encoder and visual encoder using the contrastive loss based on 400 millions image-text association pairs, which demonstrates a good generability for the unseen classes.
Similar to the previous work CoOp and CoCoOp, we apply the pretrained CLIP for knowledge transfer. 

\paragraph{Prompt Tuning:} To adapt the pretrained VLM to the downstream tasks, the prompt tuning~\cite{abs-2210-09263} always applies task-related textual tokens to infer the task-specific textual knowledge~\cite{abs-2210-07225,RadfordKHRGASAM21}.
For example, the hand-crafted template ``a photo of a [CLASS]" in CLIP~\cite{RadfordKHRGASAM21} is used to model the textual embedding for zero-shot prediction.
However, the hand-crafted prompts have less ability to describe the downstream task because they do not consider the specific knowledge of the current task.
To address the above problem, Context Optimization(CoOp)~\cite{ZhouYLL22} replaces the hand-crafted prompts with the learnable soft prompts inferred by the labeled few-shot samples.
The disadvantage of CoOp is that the learnable prompts are unique and fixed for each task's images.
That is to say, CoOP infers task-related prompts and ignores the characteristics of different images.
Furthermore, Conditional Context Optimization(CoCoOp)~\cite{ZhouYL022} is proposed to generate an image-conditional context for each image and combine the textual-conditional context for prompt tuning.
Specialy, it uses a lightweight neural network to generate a vector, which is learnable text prompts.
To obtain high-quality task-related tokens, ProDA~\cite{proda} considers the prompt's prior distribution learning.
Furthermore, ProGrad~\cite{abs-2205-14865} only updates the prompts whose gradient is aligned to the ``general knowledge" generated by the original prompts. 
DenseCLIP~\cite{RaoZ0TZH0L22} uses the context-aware prompt strategy to generate dense prediction tasks, and CLIP-Adapter~\cite{abs-2110-04544} applies an adapter to adjust the visual or text embeddings.

Among existing methods, the most related to ours are the CoOp and ProGrad.
The CoOp can be treated as the baseline model for the proposed KgCoOp.
Compared with CoOp, the proposed KgCoOp considers an additional term to ensure learnable prompts have a low discrepancy with the original prompts, leading the proposed KgCoOp to obtain a higher performance on the terms of unseen classes than CoOp.
ProGrad has the same idea as the proposed KgCoOp, ensuring that the learnable specific knowledge is aligned with the general knowledge.
However, ProGrad only optimizes the prompts with the aligned direction and discards a conflicting update.
That is to say, ProGrad discards a lot of the conflict knowledge during prompt tuning.
Unlike ProGrad, the proposed KgCoOp will not discard any knowledge and only ensures that the learnable specific knowledge is close to the general knowledge.
Furthermore, KgCoOp is more efficient than ProGrad because it does not need additional computation.
The comprehensive evaluation shows that the proposed KgCoOp is an efficient method: using less training time obtains a higher performance.

\section{Methodolgy}
As Knowledge-guided Context Optimization(KgCoOp) is proposed based on Context Optimization (CoOp), we first give a brief review of Context Optimization (CoOp) for visual-language prompt tuning.
Then, we give a detailed introduction to the proposed KgCoOp.

\subsection{ Preliminaries}
Among the existing visual-language models, Contrastive Language-Image Pre-training(CLIP) is a representative model trained with 400 million image-text association pairs, having a powerful generability for zero-shot image recognition. 
Since CLIP is trained based on the image-text association pairs,  it  contains two types of encoders: visual encoder, and textual encoder, where the visual encoder is used to map the given image into the visual embedding, and the textual encoder is applied to embedding the corresponding textual information.
By fixing the pretrained visual and textual encoders in CLIP, the prompt tuning uses the hand-crafted prompts or  the learnable prompts for adapting the pre-trained CLIP to downstream tasks.

Formally, we define the visual encoder and textual encoder as $\phi$ and $\theta$, respectively.
For a downstream task consisting of ${N}_c$ categories,  CLIP employs a hand-crafted prompt to generate the textual class embeddings, \emph{i.e.,}
$\mathbf{W}^{clip}=\{\mathbf{w}^{clip}_i\}_{i=1}^{{N}_c}$ denotes the textual embedding of all categories, where $\mathbf{w}^{clip}_i$ denotes the textual embedding of the $i$-th class.
Specifically, assuming the name of the $i$-th class as ``class-name", the corresponding textual embedding $\mathbf{w}^{clip}_i$ is generated from a hand-crafted prompt: ``a photo of a [class-name]'' with the textual encoder $\theta(\cdot)$ and a transformer-based encoder $e(\cdot)$, where $e(\cdot)$ takes a sequence of words as input and outputs a vectorized textual tokens.
Formally, the vectorized textual tokens of the $i$-th class template ``a photo of a [class-name]'' is defined as: $\mathbf{t}^{clip}_i=e$(``a photo of a [class-name]''). 
$\mathbf{t}^{clip}_i$ is further project to the textual class embedding $\mathbf{w}^{clip}_i$ with the textual encoder $\theta$: $\mathbf{w}^{clip}_i=\theta(\mathbf{t}^{clip}_i)$.

Given an image $I$ along with its label $y$,  the visual embedding is extracted with the visual encoder $\phi(\cdot)$: $\mathbf{x}=\phi(I)$.
After that, the prediction probability between the visual embedding $\mathbf{x}$ and textual embedding $\mathbf{W}^{clip}$ is computed for prediction: 

\begin{equation}
p(y|\mathbf{x})=\frac{\exp(d(\mathbf{x},\mathbf{w}^{clip}_y)/\tau)}{\sum_{i=1}^{N_c}\exp(d(\mathbf{x},\mathbf{w}^{clip}_i)/\tau)},
\label{eq:clip}
\end{equation} 
where $d(\cdot)$ denotes the cosine similarity, and $\tau$ is a learnable temperature parameter.

Although Eq.\eqref{eq:clip} can be easily applied for zero-shot prediction, CLIP employs a fixed hand-crafted prompt(``a photo of a []") to generate the textual embedding, leading to weak generability to the downstream tasks.
To address the above problem, Context Optimization (CoOp) automatically learns a set of continuous context vectors for generating task-related textual embeddings.
Specifically, CoOp introduces $M$ context vectors $\mathbb{V}=\{\mathbf{v}_1, \mathbf{v}_2,...,\mathbf{v}_{M}\}$ as the learnable prompt.
Finally, the corresponding class token embedding $\mathbf{c}_i$ of the $i$-th class is concatenated with the learnable context vector $\mathbb{V}$ for generating the prompts $\mathbf{t}^{coop}_i=\{\mathbf{v}_1, \mathbf{v}_2,...,\mathbf{v}_{M},\mathbf{c}_i\}$.
After that, the textual class embedding $\mathbf{w}^{coop}_{i}$ is obtained by fedding the learnable prompts $\mathbf{t}^{coop}_i$ into the textual encoder $\theta$, \emph{i.e.}, $\mathbf{w}^{coop}_i=\theta(\mathbf{t}^{coop}_i)$.
Therefore, the final textual class embedding for all class is defined as: $\mathbf{W}^{coop}=\{\mathbf{w}^{coop}_i\}_{i=1}^{{N}_c}$.

With the given few-shot samples, CoOp optimizes the learnable context tokens $\mathbb{V}$ by minimizing the negative log-likelihood between the image feature $\mathbf{x}$ and its class textual embeding $\mathbf{w}^{coop}_{y}$:
\begin{equation}
p_{coop}(y|\mathbf{x})=\frac{\exp(d(\mathbf{x},\mathbf{w}^{coop}_y)/\tau)}{\sum_{i=1}^{N_c}\exp(d(\mathbf{x},\mathbf{w}^{coop}_i)/\tau)}.
\label{eq:coop}
\end{equation} 

Note that the visual encoder $\phi$ and the pretrained textual encoder $\theta$ are frozen during training for CLIP and CoOp.
Different from CLIP using the fixed prompts, CoOp only infers the suitable task-related prompts $\mathbf{t}^{coop}_i$ to boost its generability and discrimination.

\subsection{Knowledge-guided Context Optimization}
\begin{figure}
  \centering
   \includegraphics[width=1.0\linewidth]{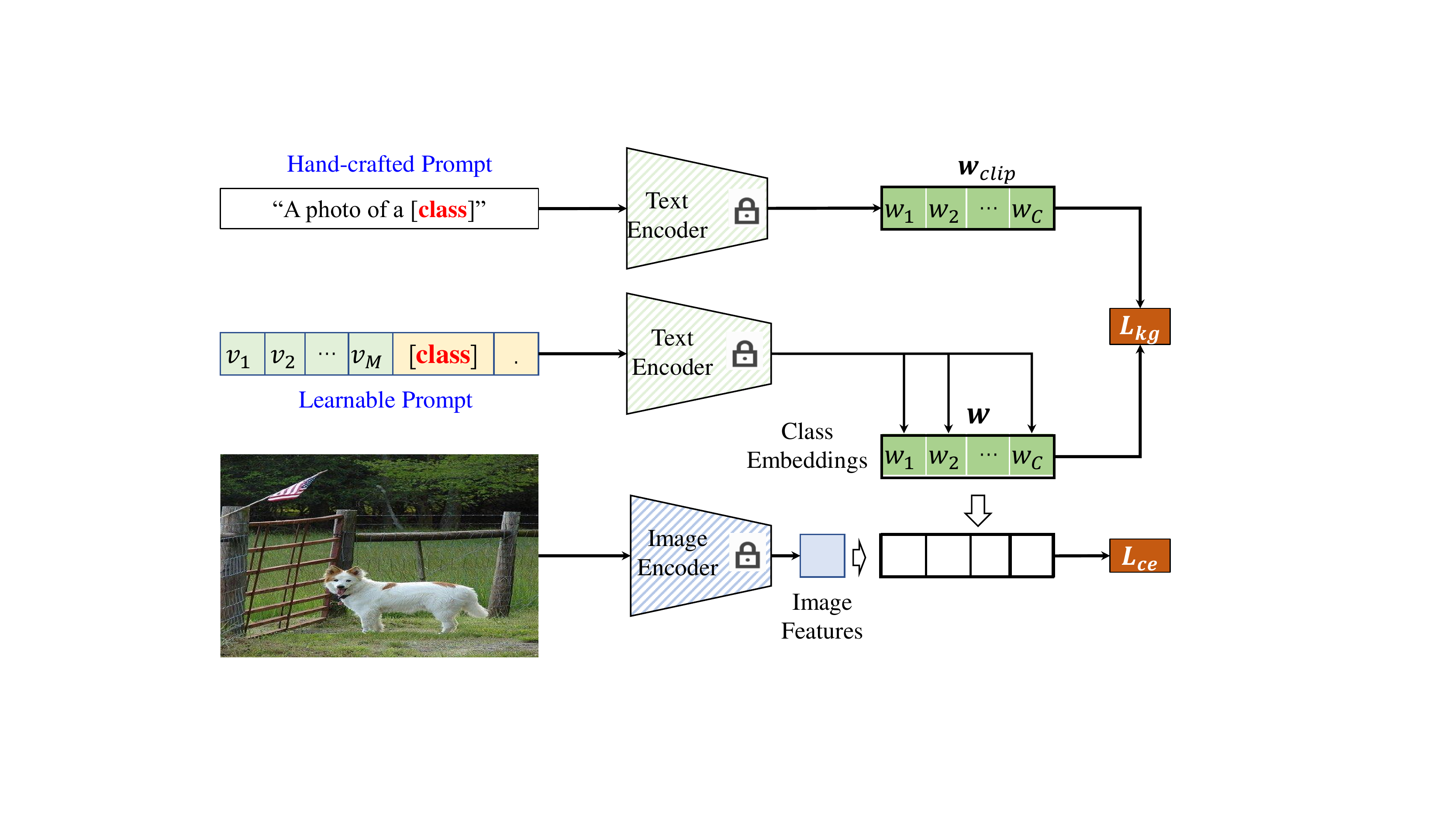}
   \caption{The framework of the Knowledge-guided Context Optimization for prompt tuning. $\mathcal{L}_{ce}$ is the standard cross-entropy loss, and $\mathcal{L}_{kg}$ is the proposed Knowledge-guided Context Optimization contraint to minimize the discrepancy  between the special knowledge (learnable textual embeddings) and the general knowledge(the textual embeddings generated by the hand-crafted prompt).}
   \label{fig:KgCoOp}
\vspace{-1.0em}
\end{figure}

Although existing CoOp-based prompt tuning methods can effectively adapt the pretrained CLIP to the downstream tasks, it might easily overfit the seen classes because only a few labeled images are used for training.
For example, CoOp obtains a noticeable improvement for the \emph{Base} accuracy upon CLIP, \emph{e.g.,} 69.34\%(CLIP) vs 82.89\%(CoOp). 
However, CoOp obtains a worse \emph{New} accuracy than CLIP on the unseen classes, \emph{e.g.,} 74.22\%(CLIP) vs 63.22\%(CoOp). 
By further analyzing the \emph{New} accuracy between CLIP and CoOp on all 11 datasets,  an interesting phenomenon is that the performance degradation on the unseen classes is consistent with the distance between the learnable prompts and fixed prompts.
In this work, the relative ratio of performance drop $\triangledown_{new}$ between CLIP and CoOp indicates the degree of performance degradation.
Moreover, the distance between learnable textual embedding (CoOp) and fixed textual embedding(CLIP) is used to measure the similarity between the two types of prompts.
As shown in Figure~\ref{fig:motivation}, the larger distance, the more severe the performance drop. 
For example, among all 11 datasets, CoOp obtains the largest drop ratio of 20.63\% on the DTD dataset, while its special class embeddings also have the largest distance compared to CLIP ones.
Based on the above results, we can conclude that enhancing the similarity between the learnable prompt and fixed prompts can alleviate the forgetting of general textual knowledge for boosting the generability of the unseen domain, which is the core motivation of our work.
Formally, we propose a novel prompt tuning method named Knowledge-guided Context Optimization (KgCoOp) to infer learnable prompts which have a high discriminative on the seen classes and high generability on the unseen classes, shown in Figure~\ref{fig:KgCoOp}. 

\begin{table*}[]
\caption{Comparison in the base-to-new setting with different $K$-shot samples in terms of the average performance among all 11 datasets and backbones(ViT-B/16 and ResNet-50).}
\label{tab:k}
\centering
\small
\vspace{-0.8em}
\begin{tabular}{l|l|ccc||ccc||ccc}
\toprule
  Backbones & Methods& \multicolumn{3}{c||}{$K$=4}                                            & \multicolumn{3}{c||}{$K$=8}                                            & \multicolumn{3}{c}{$K$=16}                                           \\ \cline{3-11} 
        & & Base                 & New                  & H                    & Base                 & New                  & H                    & Base                 & New                  & H                    \\
\midrule
& CoOp    & 78.43                & 68.03                & 72.44                & 80.73                & 68.39                & 73.5                 & 82.63                & 67.99                & 74.60                \\
& CoCoOp  & 76.72                & \textbf{73.34}                & 74.85                & 78.56                & 72.0                 & 74.9                 & 80.47                & 71.69                & 75.83                \\
ViT-B/16& ProGrad & 79.18                & 71.14                & 74.62                & \textbf{80.62}                & 71.02                & 75.2                 & \textbf{82.48}                & 70.75                & 76.16                \\
& KgCoOp     & \textbf{79.92}                & 73.11                & \textbf{75.90}                &      78.36  &    \textbf{73.89}         &     \textbf{76.06}            & 80.73                & \textbf{73.6}                 & \textbf{77.0}                 \\
\midrule
\midrule
& CoOp    & 72.06   & 59.69                & 65.29               & 74.72                & 58.05               & 65.34                & 77.24                & 57.4                & 65.86                \\
& CoCoOp  & 71.39   & 65.74                & 68.45                & 73.4               & 66.42                 & 69.29                &  75.2&          63.64   &       68.9     \\
ResNet-50& ProGrad & \textbf{73.88}               & 64.95                & 69.13                & \textbf{76.25}                & 64.74                & 70.03                 & \textbf{77.98}                & 64.41                & 69.94                \\
& KgCoOp     & 72.42             & \textbf{68.00 }               & \textbf{70.14}                & 74.08 &     \textbf{67.86} &    \textbf{70.84}  &    75.51 &  \textbf{67.53} &  \textbf{71.30}  \\
\bottomrule
\end{tabular}
\vspace{-1.0em}
\end{table*}

For CLIP, given an image $I$ along with its embedding $\mathbf{x}$, the predictions are obtained by computing the visual-textual similarity between the visual embedding and textual class embeddings.
Since CLIP and KgCoOp apply different textual embeddings to match the visual embeddings, the general textual knowledge and special textual knowledge are majorly controlled by the textual embeddings of CLIP and KgCoOp.
Furthermore, the discrepancy between general textual knowledge and special textual knowledge can be measured by the distance between the corresponding textual embeddings. 

Formally,  we define the textual embedding generated by the CLIP and KgCoOp as $\mathbf{w}^{clip}_{i}=\theta(\mathbf{t}^{clip}_i)$ and $\mathbf{w}_i=\theta(\mathbf{t}_i)$, where $\mathbf{t}^{clip}_i$ is the vectorized textual tokens in CLIP, and $\mathbf{t}_i=\{\mathbf{v}_1, \mathbf{v}_2,...,\mathbf{v}_{M},\mathbf{c}_i\}$ denotes the learnable prompt of the $i$-th class.
The discrepancy between the special knowledge and general knowledge is to compute the euclidean distance between $\mathbf{w}_i$ and $\mathbf{w}^{clip}_i$,
As shown in Figure~\ref{fig:motivation}, the distance is positively related to the performance degradation, and a lower distance produces a lower performance degradation.
Therefore, we can minimize the distance between  $\mathbf{w}_i$  and $\mathbf{w}^{clip}_i$ for boosting the generability of the unseen classes,
\begin{equation}
\mathcal{L}_{kg}=\frac{1}{N_c}\sum_{i=1}^{N_c}||\mathbf{w}_i-\mathbf{w}^{clip}_i||^{2}_{2},
\end{equation}
where $||\cdot||$ is the euclidean distance, $N_c$ is the number of seen classes. 
Meanwhile, the standard contrastive loss is:
\begin{equation}
\mathcal{L}_{ce}=-\sum_{\mathbf{x}\in \mathbf{X}} \log\frac{\exp(d(\mathbf{x},\mathbf{w}_y)/\tau)}{\sum_{i=1}^{N_c}\exp(d(\mathbf{x},\mathbf{w}_i)/\tau)},
\label{eq:ce}
\end{equation} 
where $y$ is the corresponding label of the image embedding.

By combining the standard cross-entropy loss $\mathcal{L}_{ce}$, the final objective is:
\begin{equation}
\mathcal{L} = \mathcal{L}_{ce}+\lambda\mathcal{L}_{kg}, 
\end{equation}
where $\lambda$ is used balance the effect of $\mathcal{L}_{kg}$ in the final objective.

\section{Experiments}
Similar to CoCoOp~\cite{ZhouYL022} and ProGrad~\cite{abs-2205-14865}, we evaluate the proposed method based on the following settings: 1) generalization from base-to-new classes within a dataset; 2) few-shot image classification; 3) domain generalization.
All experiments are conducted based on the pretrained CLIP~\cite{RadfordKHRGASAM21} model.
More detailed results will be given in the Supplementary materials.

\textbf{Dataset:}
Following CLIP~\cite{RadfordKHRGASAM21}, CoOp~\cite{ZhouYLL22}, CoCoOp~\cite{ZhouYL022}, and ProGrad~\cite{abs-2205-14865}, the base-to-new generaliation is conducted on 11 image classification datasets, \emph{i.e.,} ImageNet~\cite{DengDSLL009} and Caltech~\cite{Fei-FeiFP07} for generic object classification; OxfordPets~\cite{ParkhiVZJ12}, StanfordCars~\cite{Krause0DF13}, Flowers~\cite{NilsbackZ08}, Food101~\cite{BossardGG14}, and FGVCAircraft~\cite{MajiRKBV13} for fine-grained visual categorization, EuroSAT~\cite{HelberBDB19} for satellite image classification, UCF101~\cite{abs-1212-0402} for action recognization, DTD~\cite{CimpoiMKMV14} for texture classification, and SUN397~\cite{XiaoHEOT10} for scene recognition.
Furthermore, we use the ImageNet and its variants for domain generalization, \emph{i.e.,} the ImageNet is treated as the source domain; ImageNetV2~\cite{RechtRSS19}, ImageNet-Sketch~\cite{WangGLX19}, ImageNet-A~\cite{GaoZYLGW22} and ImageNet-R~\cite{HendrycksBMKWDD21} are treated as the target domains for evaluation.

\textbf{Training Details:}
Our implementation is based on CoOp's~\cite{ZhouYLL22}~\footnote{https://github.com/KaiyangZhou/CoOp} and ProGrad's~\cite{abs-2205-14865}~\footnote{https://github.com/BeierZhu/Prompt-align} codes with the CLIP model.
We conduct the experiments based on the vision backbone with ResNet-50~\cite{HeZRS16} and Vit-B/16~\cite{DosovitskiyB0WZ21}.
Inspired by CoOp, we fix the context length to 4 and initialize the context vectors using the template of ``a photo of a []".
The final performance is averaged over three random seeds for a fair comparison.
We follow the same training epochs, training schedule, and data augmentation setting in CoOp and ProGrad.
The hyperparameter $\lambda$ is set to 8.0.
All experiments are conducted based on RTX 3090.

\textbf{Baselines:}
 Four type of CoOp-based methods are treated as baselines for comparison:
\begin{itemize}[itemsep=2pt,topsep=0pt,parsep=0pt]
\item CLIP~\cite{RadfordKHRGASAM21} applies the hand-crafted template ``a photo of a []" to generate the prompts for knowledge transfer.
\item CoOp~\cite{ZhouYLL22} replaces the hand-crafted prompts with a set of learnable prompts inferred by the downstream datasets, which is our baseline.
\item CoCoOp~\cite{ZhouYL022} generates the image-conditional prompts by combining the image context of each image and the learnable prompts in CoOp.
\item ProGrad~\cite{abs-2205-14865} uses the same prompts as CoOp while only optimizing the prompt whose gradient is aligned to the ``general direction", which can be treated as CoOp+Grad.
\item KgCoOp uses the same prompts as CoOp while optimizing the learnable prompts closed to the fixed prompts in CLIP, which can be treated as CoOp+Kg.
\end{itemize}

Although the existing VPT~\cite{abs-2210-02390} and ProDA~\cite{proda} have been proposed for prompt tuning, they both infer a collection of prompts rather than one learnable prompt used in CoOp-based methods.

\begin{table*}
\caption{Comparison with existing methods in the base-to-new generalization setting with ViT-B/16 as the backbone. 
The context length $M$ is 4 for prompot-based methods with the 16-shots samples from the base classes. H: Harmonic mean.}
\label{tab:main}
\vspace{-0.8em}
\begin{minipage}{0.32\textwidth}
\centering
\footnotesize
 \subcaption{Average over 11 datasets.}
\begin{tabular}{l|cc|c}
\toprule
	&Base & New & H\\
\midrule
CLIP 		& 69.34 & \textbf{74.22} & 71.70\\
CoOp		& \textbf{82.63} & 67.99 & 74.60\\
CoCoOp 	& 80.47	& 71.69 & 75.83\\
ProGrad 	& 82.48& 70.75 & 76.16\\
\midrule
KgCoOp		&80.73	& 73.6	& \textbf{77.0}\\
\bottomrule
\end{tabular}
\end{minipage}
\hfill
\begin{minipage}{0.32\textwidth}
\centering
\footnotesize
 \subcaption{ImageNet.}
\begin{tabular}{l|cc|c}
\toprule
	&Base & New & H\\
\midrule
CLIP 		&72.43 & 68.14 & 70.22\\
CoOp		&76.46 & 66.31 & 71.02\\
CoCoOp 	&75.98 & \textbf{70.43} & \textbf{73.10} \\
ProGrad 	&\textbf{77.02}	& 66.66 & 71.46\\
\midrule
KgCoOp		&75.83 & 69.96	& 72.78\\
\bottomrule
\end{tabular}
\end{minipage}
\hfill
\begin{minipage}{0.32\textwidth}
\centering
\footnotesize
\subcaption{Caltech101.}
\begin{tabular}{l|cc|c}
\toprule
	&Base & New & H\\
\midrule
CLIP 		&96.84 &94.00 &95.40 \\
CoOp		&\textbf{98.11} &93.52 &95.76\\
CoCoOp 	&97.96 &93.81 &95.84 \\
ProGrad 	&98.02 & 93.89 & 95.91\\
\midrule
KgCoOp		&97.72 & \textbf{94.39} & \textbf{96.03}\\
\bottomrule
\end{tabular}
\end{minipage}
\hfill
\begin{minipage}{0.32\textwidth}
\centering
\footnotesize
 \subcaption{OxfordPets.}
\begin{tabular}{l|cc|c}
\toprule
	&Base & New & H\\
\midrule
CLIP 		&91.17 &97.26 &94.12\\
CoOp		&94.24&96.66	&95.43\\
CoCoOp 	&\textbf{95.20} &97.69	&\textbf{96.43} \\
ProGrad 	&95.07	&97.63 & 96.33 \\
\midrule
KgCoOp		&94.65&\textbf{97.76}&96.18\\
\bottomrule
\end{tabular}
\end{minipage}
\hfill
\begin{minipage}{0.32\textwidth}
\centering
\footnotesize
 \subcaption{StanfordCars.}
\begin{tabular}{l|cc|c}
\toprule
	&Base & New & H\\
\midrule
CLIP 		&63.37 &74.89 &68.65\\
CoOp		&76.2&69.14 &72.49\\
CoCoOp 	&70.49 &73.59 &72.01\\
ProGrad 	&\textbf{77.68} &68.63 &72.88\\
\midrule
KgCoOp		&71.76&\textbf{75.04}& \textbf{73.36}\\
\bottomrule
\end{tabular}
\end{minipage}
\hfill
\begin{minipage}{0.32\textwidth}
\centering
\footnotesize
\subcaption{Flowers102.}
\begin{tabular}{l|cc|c}
\toprule
	&Base & New & H\\
\midrule
CLIP 		&72.08 &\textbf{77.80} &74.83 \\
CoOp		&\textbf{97.63} &69.55 & 81.23\\
CoCoOp 	&94.87&71.75&81.71\\
ProGrad 	&95.54&71.87&82.03\\
\midrule
KgCoOp		&95.00&74.73&\textbf{83.65}\\
\bottomrule
\end{tabular}
\end{minipage}
\hfill
\begin{minipage}{0.32\textwidth}
\centering
\footnotesize
 \subcaption{Food101.}
\begin{tabular}{l|cc|c}
\toprule
	&Base & New & H\\
\midrule
CLIP 		&90.10&91.22&90.66	\\
CoOp		&89.44&87.50&88.46	\\
CoCoOp 	&\textbf{90.70}&91.29&90.99\\
ProGrad 	&90.37&89.59& 89.98\\
\midrule
KgCoOp		&90.5&\textbf{91.7}&\textbf{91.09}\\
\bottomrule
\end{tabular}
\end{minipage}
\hfill
\begin{minipage}{0.32\textwidth}
\centering
\footnotesize
 \subcaption{FGVCAircraft.}
\begin{tabular}{l|cc|c}
\toprule
	&Base & New & H\\
\midrule
CLIP 		&27.19&\textbf{36.29}&31.09	\\
CoOp		&39.24&30.49&34.30	\\
CoCoOp 	&33.41&23.71&	27.74\\
ProGrad 	&\textbf{40.54}&27.57& 32.82\\
\midrule
KgCoOp		&36.21&33.55&\textbf{34.83}\\
\bottomrule
\end{tabular}
\end{minipage}
\hfill
\begin{minipage}{0.32\textwidth}
\centering
\footnotesize
\subcaption{SUN397.}
\begin{tabular}{l|cc|c}
\toprule
	&Base & New & H\\
\midrule
CLIP 		&69.36&75.35&	72.23\\
CoOp		&80.85&68.34&74.07\\
CoCoOp 	&79.74&\textbf{76.86}&78.27\\
ProGrad 	&\textbf{81.26}&74.17& 77.55\\
\midrule
KgCoOp		&80.29& 76.53& \textbf{78.36}\\
\bottomrule
\end{tabular}
\end{minipage}
\hfill
\begin{minipage}{0.32\textwidth}
\centering
\footnotesize
 \subcaption{DTD.}
\begin{tabular}{l|cc|c}
\toprule
	&Base & New & H\\
\midrule
CLIP 		&53.24&\textbf{59.90}&56.37\\
CoOp		&\textbf{80.17}&47.54&59.68\\
CoCoOp 	&77.01&56.00&\textbf{64.85}\\
ProGrad 	&77.35&52.35&62.45\\
\midrule
KgCoOp		&77.55& 54.99& 64.35\\
\bottomrule
\end{tabular}
\end{minipage}
\hfill
\begin{minipage}{0.32\textwidth}
\centering
\footnotesize
 \subcaption{EuroSAT.}
\begin{tabular}{l|cc|c}
\toprule
	&Base & New & H\\
\midrule
CLIP 		&56.48&64.05&60.03\\
CoOp		&\textbf{91.54}&54.44&68.27\\
CoCoOp 	&87.49&60.04&71.21	\\
ProGrad 	&90.11&60.89&72.67\\
\midrule
KgCoOp		&85.64& \textbf{64.34}& \textbf{73.48}\\
\bottomrule
\end{tabular}
\end{minipage}
\hfill
\begin{minipage}{0.32\textwidth}
\centering
\footnotesize
\subcaption{UCF101.}
\begin{tabular}{l|cc|c}
\toprule
	&Base & New & H\\
\midrule
CLIP 		&70.53&\textbf{77.50}&73.85\\
CoOp		&\textbf{85.14}&64.47&	73.37\\
CoCoOp 	&82.33&73.45&77.64	\\
ProGrad 	&84.33&74.94&79.35\\
\midrule
KgCoOp		&82.89& 76.67& \textbf{79.65}\\
\bottomrule
\end{tabular}
\end{minipage}
\vspace{-0.8em}
\end{table*}

\subsection{Generalization From Base-to-New Classes}
Similar to the previous work CoOp and CoCoop, we split each dataset into two groups: base classes (\emph{Base}) and new classes(\emph{New}).
Similar to the zero-shot setting, the new classes disjoint the base classes.
To verify the generalization of the CoOp-based methods, all compared methods and the proposed KgCoOp use the base classes for prompt tuning, and conduct evaluation on the new class.
The detailed results are shown in Table~\ref{tab:k} and Table~\ref{tab:main}.
Table~\ref{tab:k} summarizes the average performance among all 11 datasets with different $K$-shot samples and backbones (ViT-B/16 and ResNet-50).
Table~\ref{tab:main} gives the detailed performance on all 11 datasets based on the backbone of ViT-B/16 and 16-shot samples.

\textbf{Total Analysis:} As shown in Table~\ref{tab:k}, the proposed \emph{KgCoOp} obtains a higher average performance in terms of Harmonic mean than existing methods on all settings, demonstrating its superiority for the generalization from base-to-new classes.
Among the existing methods, ProGrad obtains the best performance in terms of \emph{Base} classes on all settings while obtaining a worse \emph{New} performance than CoCoOp.
The reason is that a higher performance on \emph{Base} classes makes the ProGrad have serious overfitting on the \emph{Base} class, thus producing a biased prompt for the \emph{New} classes, leading to a worse \emph{New} performance.
Compared with CoCoOp, the proposed KgCoOp slightly improves the \emph{Base} classes.
For example, based on the backbone of ViT-B/16, KgCoOp achieves the \emph{Base} performance of 78.36\% and 80.73\% for the 8-shot and 16-shot settings respectively, which are similar to the 78.56\% and 80.47\% obtained by CoCoOp.
However, KgCoOp obtains a significant improvement on the \emph{New} class  upon CoCoOp, \emph{e.g.,} obtains the improvement of 1.89\% and 1.91\% upon CoCoOp for 8-shot and 16-shot setting, respectively.
The superior performance on \emph{New} classes demonstrates that the KgCoOp can improve the generability of the wider unseen class without discarding the discriminative ability of the seen classes.

\begin{table*}[]
\centering
\small
\caption{Comparison of prompt learning in the domain generalization with 16-shot source samples. where ``vp" and ``tp" denote the visual prompting and textual prompting, respectively.}
\label{tab:dg}
\vspace{-0.8em}
\begin{tabular}{l|c|c|cccc|c}
\toprule
        &Prompts& Source   & \multicolumn{5}{c}{Target}                             \\\cline{3-8} 
        & & ImageNet & ImageNetV2 & ImageNet-Sketch & ImageNet-A & ImageNet-R & Avg. \\
\midrule
CLIP~\cite{RadfordKHRGASAM21} &  hand-crafted	& 66.73    	& 60.83    	& 46.15          	& 47.77      	& 73.96      & 57.17\\ 
UPT~\cite{abs-2210-07225} & vp+tp& \textbf{72.63} & 64.35 & 48.66 & 50.66 & 76.24 & 59.98 \\
CoCoOp~\cite{ZhouYL022}& vp+tp  	& 71.02    	& 64.07      	& 48.75           	& 50.63      	& 76.18      & 59.90 \\
CoOp~\cite{ZhouYLL22}&  tp  	& 71.51    	& 64.2       	& 47.99           	& 49.71      	& 75.21    & 59.28   \\
ProGrad~\cite{abs-2205-14865}& tp 	& 72.24    	& \textbf{64.73}      	& 47.61           	& 49.39      	& 74.58   & 59.07   \\
KgCoOp  &  tp & 71.2    	& 64.1       	& \textbf{48.97}           	& \textbf{50.69}     	& \textbf{76.7}  & \textbf{60.11}  \\
\bottomrule
\end{tabular}
\vspace{-0.8em}
\end{table*} 

As mentioned above, ProGrad obtains a better performance on the \emph{Base} class and a worse performance on the \emph{New} classes, leading to the generated prompt having serious overfitting on the \emph{Base} classes.
Since KgCoOp aims to improve the generability of the \emph{New} class, KgCoOp also obtains a worse performance than ProGrad on the term of \emph{Base} classes.
However, KgCoOp obtains a higher performance on the \emph{New} class.
By improving the generability of \emph{New} class, KgCoOp obtains a higher performance in terms of $H$ than ProGrad, \emph{e.g.,} improving the harmonic mean (H) from 75.2\% and 76.16\% to 76.06\% and 77.0\% for the 8-shot and 16-shot settings, respectively.
The superior performance demonstrates that the KgCoOp can effectively adapt the pretrained VLM model on the downstream task with improving the generality of the unseen classes.

\begin{table}[]
\caption{Accuracy (\%) of few-shot(K=4) learning on 11 datasets.}
\label{tab:fsl}
\centering
\small
\vspace{-1.0em}
\begin{tabular}{l|ccc|c}
\toprule
Datasets & CoOp & CoCoOp & ProGrad & KgCoOp \\
\midrule
ImageNet & 69.38 & \textbf{70.55} & 70.21 & 70.19 \\
Caltech101 & 94.44 &  \textbf{94.98}& 94.93 & 94.65 \\
OxfordPets & 91.3 &    \textbf{93.01}& 93.21 & 93.2 \\
StanfordCars & \textbf{72.73} &   69.1& 71.75 & 71.98 \\
Flowers102 & \textbf{91.14} &  82.56& 89.98 & 90.69 \\
Food101 & 82.58 &   \textbf{86.64}& 85.77 & 86.59 \\
FGVCAircraft & 33.18 &  30.87 & \textbf{32.93} & 32.47 \\
SUN397 & 70.13 & 70.5& 71.17 & \textbf{71.79} \\
DTD & \textbf{58.57} &54.79& 57.72 & 58.31 \\
EuroSAT & 68.62 &63.83& 70.84 & \textbf{71.06} \\
UCF101 & 77.41 &  74.99& 77.82 & \textbf{78.40} \\
\midrule
Avg. & 73.59 & 71.98 & 74.21 & \textbf{74.48} \\
\bottomrule
\end{tabular}
\vspace{-1.0em}
\end{table}

\textbf{Detailed Analysis:} We thus give a detailed comparison of each dataset for the prompt-based method with a 16-shot setting with the ViT-B/16 as the backbone.
As shown in Table~\ref{tab:main},  existing CoOp-based methods, \emph{i.e.,} CoOp, CoCoOp, and ProGrad, all significantly improve the \emph{Base} classes compared to CLIP on all 11 datasets.
Especially, ProGrad, CoOp, and CoCoOp obtain the best \emph{Base} performance on 4/11 datasets, 5/11 datasets, and 2/11 datasets, respectively.
While the CoOp also obtains a better average \emph{Base} performance than ProGrad and CoCoOp.
The reason is that CoOp only focuses on inferring a learnable prompt without considering any other constraints, making the generated prompt be discriminative for the \emph{Base} class.
Unlike CoOp, CoCoOp considers the instance-conditional token combined with the learnable context vectors.
Using the instance-conditional token can improve the generability on the \emph{New} class, while degrading the discrimination on the \emph{Base} class.
Therefore, CoCoOp obtains the best \emph{New} performance on 6/11 datasets, and the best average \emph{New} performance.
Specially, CoCoOp obtains an obivous performance improvement of 3.77\%, 4.96\%, 1.7\% and 3.65\% on ImageNet, StandfordCars, Food101, and DTD upon ProGrad, respectively,
while ProGrad obtains the obvious performance improvement upon CoCoOp for the FGVCAircraft datasets, \emph{e.g.,} 23.71\%(CoCoOp) \emph{vs} 27.57\%(ProGrad).
However, existing methods, \emph{i.e.,} CoOp, CoCoOp, and ProGrad, all obtain a worse performance than the original CLIP in most cases, which indicates that they weaken generability to the \emph{New} classes.
Compared with existing methods, the proposed KgCoOp obtains a higher \emph{New} performance on eight datasets among all 11 datasets, \emph{e.g.,} Caltech101, OxfordPets, StanfordCars, Flowers102, Food101, FGVCAircraft, EuroSAT, and UCF101.
The superior performance demonstrates that KgCoOp has a better generability to the \emph{New} classes than existing CoOp-based prompt methods.
Meanwhile, in most cases, KgCoOp obtains the same performance as CoCoOp on the \emph{Base} classes.
Therefore, KgCoOp can improve the generability on \emph{New} classes without degrading the performance of \emph{Base} classes, leading to the best Harmonic mean on all 11 datasets.

\subsection{Domain Generalization}
Domain Generalization aims to evaluate the generalization by evaluating the trained model on the target dataset, which has the same class but different data distribution from the source domain.
Similar to CoCoOp and ProGrad, we conduct the prompt tuning on the few-shot ImageNets, and evaluate the model on the ImageNetV2, ImageNet-Sketch, ImageNet-A, and ImageNet-R.
The related results are summarized in Table~\ref{tab:dg}.

From Table~\ref{tab:dg}, we can observe that ProGrad obtains the best performance on the source ImageNet.
The superior performance shows that ProGrad can produce a discriminative prompt for the base class, consistent with the conclusion obtained in the base-to-new setting.
Similar to the comparison in the base-to-new setting, ProGrad has a weakened generability to the wider unseen classes, \emph{e.g.,} except for the ImageNetV2, ProGrad has achieved weaker performance than CoCoOp on the other three datasets and the mean performance.
Among existing methods, CoCoOp is more domain-generalizable than CoOp and ProGrad.
Compared with CoCoOp, the proposed KgCoOp obtains a higher performance on the source and target datasets, \emph{e.g.,} improving the average target performance from 59.90\% to 60.11\%.
The above comparison confirms that the learnable prompts in KgCoOp are better domain-generalizable.


\subsection{Few-shot Classification}
The base-to-new setting assumes that the new classes have different categories from the base classes, which can demonstrate the generability of different classes.
To further show the generability of the proposed method, we conduct the few-shot classification, which trains the model based on the few-shot labeled images and evaluates the model on the dataset with the same categories as the training classes.
The 4-shot setting results are summarized in Table~\ref{tab:fsl}.
We can observe that the proposed KgCoOp obtains a higher average performance than existing methods, \emph{i.e.,} CoOp, CoCoOp, and ProGad.

\subsection{Analysis}
\begin{figure}
  \centering
\vspace{-1.0em}
   \includegraphics[width=0.75\linewidth]{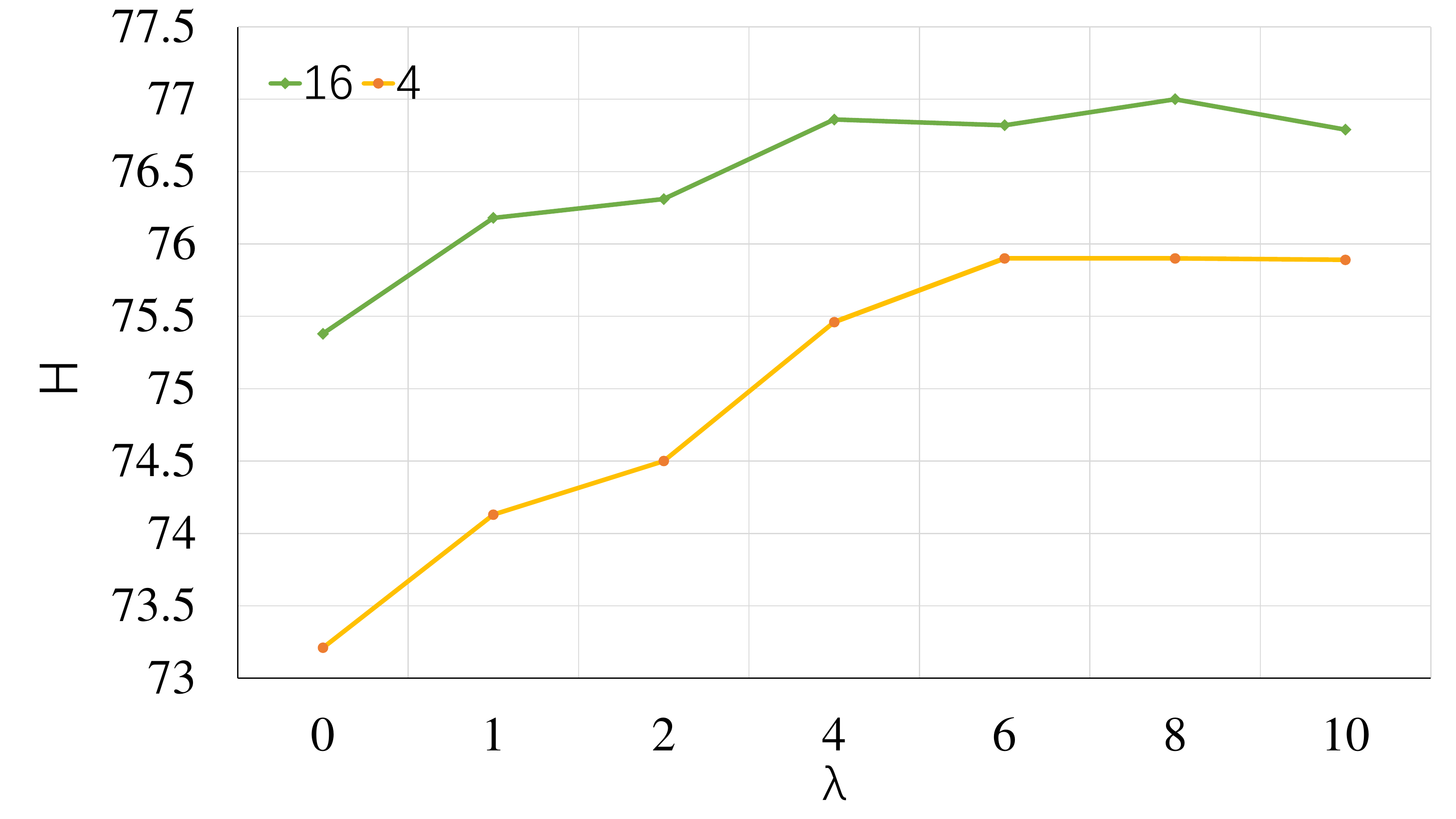}
   \caption{Effect of $\lambda$ for 4-shot and 16-shot settings on the base-to-new generalization. H: Harmonic mean}
   \label{tab:lambda}
\vspace{-0.8em}
\end{figure}

\textbf{Hyperparameter $\lambda$:}
The critical contribution of the proposed KgCoOp is applying a regularization term to constrain the special knowledge generated by prompt tuning to be closed to the general knowledge, which can improve the generalization on the unseen domain.
$\lambda$ is thus applied to balance the importance of the regularization term during prompt tuning, \emph{e.g.,} the higher $\lambda$ denotes that the prompt tuning pays more attention to the general knowledge.
We thus analyze the effect of $\lambda$, and show the results in Figure~\ref{tab:lambda}.
We can observe that a higher $\lambda$ can obtain a higher metric of $H$.
For example, setting $\lambda$ as 8.0 obtains the best performance of 77.0\%.
By further increasing $\lambda$, the performance would be degraded, \emph{e.g.,} setting $\lambda$=10.0 obtains a harmonic mean of 76.79\%, which is lower than 77.0\% for $\lambda$=8.0.

\textbf{Effect of $\mathcal{L}_{kg}$}:
The critical of our work is to use the constraint $\mathcal{L}_{kg}$ to  minimize the general textual embedding and specific textual embedding, which can be easily applied to existing CoOp-based methods, \emph{e.g.,} CoOp, CoCoOp, and ProGrad.
As shown in Table~\ref{tab:lkg}, compared with CoCoOp and ProGrad, considering the additional $\mathcal{L}_{kg}$ constraint improves the performance in terms of \emph{New} and \emph{H}.
Especially for the \emph{New} performance, using  $\mathcal{L}_{kg}$  achieves more than 3\% improvement.
The superior performance further proves the effectiveness of  considering the constraint $\mathcal{L}_{kg}$ for prompt tuning.
\begin{table}
\caption{Effect of $\mathcal{L}_{kg}$ on CoOp, CoCoOp, and ProGrad in the base-to-new generalization setting with 16-shot samples and ViT-B/16 in terms of the average performance among all 11 datasets.}
\centering
\footnotesize
\label{tab:lkg}
\vspace{-0.8em}
\begin{tabular}{l|ccc}
\toprule
Methods & \emph{Base} & \emph{New} & \emph{H} \\
\midrule
\midrule
CoOp & 82.63 & 67.99 & 74.6 \\
CoOp+$\mathcal{L}_{kg}$ & 80.73$(\downarrow \textcolor{green}{-1.9})$ & 73.6$(\uparrow \textcolor{red}{5.61})$ & 77 $(\uparrow \textcolor{red}{2.4})$ \\
\midrule
\midrule
CoCoOp & 80.47 & 71.69 & 75.83 \\
CoCoOp +$\mathcal{L}_{kg}$ & 77.96 $(\downarrow \textcolor{green}{-2.50})$& 74.75$(\uparrow \textcolor{red}{3.06})$ & 76.32 $(\uparrow \textcolor{red}{0.49})$\\
\midrule
\midrule
ProGrad & 82.48 & 70.75 & 76.16 \\
ProGrad+$\mathcal{L}_{kg}$& 78.64 $(\downarrow \textcolor{green}{-3.84})$& 74.72$(\uparrow \textcolor{red}{3.97})$& 76.63 $(\uparrow \textcolor{red}{0.47})$\\
\bottomrule
\end{tabular}
\vspace{-1.5em}
\end{table}

\textbf{Quantitative analysis of  $\mathcal{L}_{kg}$:} 
KgCoOp aims to improve the generability of the unseen class by minimizing the distance $\mathcal{L}_{kg}$ between the learnable textual embedding $\mathbf{w}$ and fixed textual embedding$\mathbf{w}_{clip}$.
We thus verify the rationality and effectiveness of this motivation and summarize the related results in Table~\ref{tab:distance}.
We can observe that a higher $\lambda$ obtains a lower $\mathcal{L}_{kg}$.
Furthermore, the lower $\mathcal{L}_{kg}$, the higher performance $H$.
Therefore, we can conclude that minimizing the distance between the learnable textual embedding $\mathbf{w}$ and fixed textual embedding $\mathbf{w}^{clip}$ can improve the performance.

\textbf{Training efficienty:}
For the prompt-based method, we calculate the training time on ImageNet datasets with a16-shot setting.
Note that the batchsize is 32 for CoOp, ProGrad, and KgCoOp, while CoCoOp uses the batchsize of 1.
The training time is the average time to process one image, \emph{i.e.,} \emph{ms}/image.
Based on CoOp, the proposed KgCoOp conducts an additional constraint between the $\mathbf{w}$ and $\mathbf{w}_{clip}$ during training.
Since $\mathbf{w}_{clip}$ is a pre-computed vector generated by CLIP with the given categories names, the core of KgCoOp is merely to minimize the distance $\mathbf{w}$ and $\mathbf{w}_{clip}$.
Compared to the training time, the additional running time of the proposed method can be ignored.
As shown in Table~\ref{tab:time}, KgCoOp has the same training time as the CoOp, which is faster than CoCoOp and ProGrad.
Moreover, KgCoOp obtains the best performance.
In conclusion, KgCoOp is an efficient model achieving better performance with less training time.

\begin{table}
\caption{The quantitative analysis of  $\mathcal{L}_{kg}$ for different $\lambda$ on ImageNet.}
\label{tab:distance}
\footnotesize
\vspace{-1.0em}
\begin{tabular}{l|ccccccc}
\toprule
$\lambda$	& 0.0 & 1.0 & 2.0 & 4.0 & 6.0 & 8.0 & 10.0 \\
\midrule
$\mathcal{L}_{kg}$ & 0.18 & 0.038 & 0.024 & 0.015 & 0.010 & 0.006 & 0.005\\
$H$	& 75.38 & 76.18 & 76.31 & 76.86 & 76.82 & 77 & 76.79 \\
\bottomrule
\end{tabular}
\vspace{-0.8em}
\end{table}

\begin{table}
\caption{Traning teim comparison(\emph{ms}/image). The training time is the average time to process one image, \emph{i.e.,} \emph{ms}/image.}
\label{tab:time}
\centering
\small
\vspace{-1.0em}
\begin{tabular}{l|cccc}
\toprule
	& CoOp & CoCoOp & ProGrad & KgCoOp \\
\midrule
\emph{time}& ~6\emph{ms} & ~160\emph{ms} & ~22\emph{ms} & ~6\emph{ms} \\
\emph{H} & 74.60& 75.83& 76.16&77.0\\
\bottomrule
\end{tabular}
\vspace{-1.5em}
\end{table} 

\section{Conclusion}
To overcome the shortcoming that existing CoOp-based prompt tuning methods weaken the generability of the unseen classes, we introduce a prompt tuning method named Knowledge-guided Context Optimization to boost the generability of the unseen classes by minimizing the discrepancy between the general textual embeddings and the learnable specific textual embeddings.
Extensive evaluation of several benchmarks shows that the proposed KgCoOp is an efficient prompt tuning method.

Although using KgCoOp can improve the generability on unseen classes, it may degrade the discriminative ability on the seen class, \emph{e.g.,} KgCoOp obtains a badly \emph{Base} performance on seen classes.
We will investigate an effective method for seen and unseen classes in the future.

\newpage
{\small
\bibliographystyle{ieee_fullname}
\bibliography{egbib}

\begin{thebibliography}{10}\itemsep=-1pt

\bibitem{abs-2204-14198}
Jean{-}Baptiste Alayrac, Jeff Donahue, Pauline Luc, Antoine Miech, Iain Barr,
  Yana Hasson, Karel Lenc, Arthur Mensch, Katie Millican, Malcolm Reynolds,
  Roman Ring, Eliza Rutherford, Serkan Cabi, Tengda Han, Zhitao Gong, Sina
  Samangooei, Marianne Monteiro, Jacob Menick, Sebastian Borgeaud, Andrew
  Brock, Aida Nematzadeh, Sahand Sharifzadeh, Mikolaj Binkowski, Ricardo
  Barreira, Oriol Vinyals, Andrew Zisserman, and Karen Simonyan.
\newblock Flamingo: a visual language model for few-shot learning.
\newblock {\em CoRR}, abs/2204.14198, 2022.

\bibitem{BossardGG14}
Lukas Bossard, Matthieu Guillaumin, and Luc~Van Gool.
\newblock Food-101 - mining discriminative components with random forests.
\newblock In David~J. Fleet, Tom{\'{a}}s Pajdla, Bernt Schiele, and Tinne
  Tuytelaars, editors, {\em Computer Vision - {ECCV} 2014 - 13th European
  Conference, Zurich, Switzerland, September 6-12, 2014, Proceedings, Part
  {VI}}, volume 8694 of {\em Lecture Notes in Computer Science}, pages
  446--461. Springer, 2014.

\bibitem{ChenK0H20}
Ting Chen, Simon Kornblith, Mohammad Norouzi, and Geoffrey~E. Hinton.
\newblock A simple framework for contrastive learning of visual
  representations.
\newblock In {\em Proceedings of the 37th International Conference on Machine
  Learning, {ICML} 2020, 13-18 July 2020, Virtual Event}, volume 119 of {\em
  Proceedings of Machine Learning Research}, pages 1597--1607. {PMLR}, 2020.

\bibitem{ChoLTB21}
Jaemin Cho, Jie Lei, Hao Tan, and Mohit Bansal.
\newblock Unifying vision-and-language tasks via text generation.
\newblock In Marina Meila and Tong Zhang, editors, {\em Proceedings of the 38th
  International Conference on Machine Learning, {ICML} 2021, 18-24 July 2021,
  Virtual Event}, volume 139 of {\em Proceedings of Machine Learning Research},
  pages 1931--1942. {PMLR}, 2021.

\bibitem{CimpoiMKMV14}
Mircea Cimpoi, Subhransu Maji, Iasonas Kokkinos, Sammy Mohamed, and Andrea
  Vedaldi.
\newblock Describing textures in the wild.
\newblock In {\em 2014 {IEEE} Conference on Computer Vision and Pattern
  Recognition, {CVPR} 2014, Columbus, OH, USA, June 23-28, 2014}, pages
  3606--3613. {IEEE} Computer Society, 2014.

\bibitem{DengDSLL009}
Jia Deng, Wei Dong, Richard Socher, Li{-}Jia Li, Kai Li, and Li Fei{-}Fei.
\newblock Imagenet: {A} large-scale hierarchical image database.
\newblock In {\em 2009 {IEEE} Computer Society Conference on Computer Vision
  and Pattern Recognition {(CVPR} 2009), 20-25 June 2009, Miami, Florida,
  {USA}}, pages 248--255. {IEEE} Computer Society, 2009.

\bibitem{abs-2210-02390}
Mohammad~Mahdi Derakhshani, Enrique Sanchez, Adrian Bulat, Victor
  Guilherme~Turrisi da Costa, Cees G.~M. Snoek, Georgios Tzimiropoulos, and
  Brais Mart{\'{\i}}nez.
\newblock Variational prompt tuning improves generalization of vision-language
  models.
\newblock {\em CoRR}, abs/2210.02390, 2022.

\bibitem{DosovitskiyB0WZ21}
Alexey Dosovitskiy, Lucas Beyer, Alexander Kolesnikov, Dirk Weissenborn,
  Xiaohua Zhai, Thomas Unterthiner, Mostafa Dehghani, Matthias Minderer, Georg
  Heigold, Sylvain Gelly, Jakob Uszkoreit, and Neil Houlsby.
\newblock An image is worth 16x16 words: Transformers for image recognition at
  scale.
\newblock In {\em 9th International Conference on Learning Representations,
  {ICLR} 2021, Virtual Event, Austria, May 3-7, 2021}. OpenReview.net, 2021.

\bibitem{Fei-FeiFP07}
Li Fei{-}Fei, Robert Fergus, and Pietro Perona.
\newblock Learning generative visual models from few training examples: An
  incremental bayesian approach tested on 101 object categories.
\newblock {\em Comput. Vis. Image Underst.}, 106(1):59--70, 2007.

\bibitem{abs-2210-09263}
Zhe Gan, Linjie Li, Chunyuan Li, Lijuan Wang, Zicheng Liu, and Jianfeng Gao.
\newblock Vision-language pre-training: Basics, recent advances, and future
  trends.
\newblock {\em CoRR}, abs/2210.09263, 2022.

\bibitem{GaoZYLGW22}
Haoran Gao, Hua Zhang, Xingguo Yang, Wenmin Li, Fei Gao, and Qiaoyan Wen.
\newblock Generating natural adversarial examples with universal perturbations
  for text classification.
\newblock {\em Neurocomputing}, 471:175--182, 2022.

\bibitem{abs-2110-04544}
Peng Gao, Shijie Geng, Renrui Zhang, Teli Ma, Rongyao Fang, Yongfeng Zhang,
  Hongsheng Li, and Yu Qiao.
\newblock Clip-adapter: Better vision-language models with feature adapters.
\newblock {\em CoRR}, abs/2110.04544, 2021.

\bibitem{HeCXLDG22}
Kaiming He, Xinlei Chen, Saining Xie, Yanghao Li, Piotr Doll{\'{a}}r, and
  Ross~B. Girshick.
\newblock Masked autoencoders are scalable vision learners.
\newblock In {\em {IEEE/CVF} Conference on Computer Vision and Pattern
  Recognition, {CVPR} 2022, New Orleans, LA, USA, June 18-24, 2022}, pages
  15979--15988. {IEEE}, 2022.

\bibitem{HeZRS16}
Kaiming He, Xiangyu Zhang, Shaoqing Ren, and Jian Sun.
\newblock Deep residual learning for image recognition.
\newblock In {\em 2016 {IEEE} Conference on Computer Vision and Pattern
  Recognition, {CVPR} 2016, Las Vegas, NV, USA, June 27-30, 2016}, pages
  770--778. {IEEE} Computer Society, 2016.

\bibitem{HelberBDB19}
Patrick Helber, Benjamin Bischke, Andreas Dengel, and Damian Borth.
\newblock Eurosat: {A} novel dataset and deep learning benchmark for land use
  and land cover classification.
\newblock {\em {IEEE} J. Sel. Top. Appl. Earth Obs. Remote. Sens.},
  12(7):2217--2226, 2019.

\bibitem{HendrycksBMKWDD21}
Dan Hendrycks, Steven Basart, Norman Mu, Saurav Kadavath, Frank Wang, Evan
  Dorundo, Rahul Desai, Tyler Zhu, Samyak Parajuli, Mike Guo, Dawn Song, Jacob
  Steinhardt, and Justin Gilmer.
\newblock The many faces of robustness: {A} critical analysis of
  out-of-distribution generalization.
\newblock In {\em 2021 {IEEE/CVF} International Conference on Computer Vision,
  {ICCV} 2021, Montreal, QC, Canada, October 10-17, 2021}, pages 8320--8329.
  {IEEE}, 2021.

\bibitem{HintonVD15}
Geoffrey~E. Hinton, Oriol Vinyals, and Jeffrey Dean.
\newblock Distilling the knowledge in a neural network.
\newblock {\em CoRR}, abs/1503.02531, 2015.

\bibitem{JiaYXCPPLSLD21}
Chao Jia, Yinfei Yang, Ye Xia, Yi{-}Ting Chen, Zarana Parekh, Hieu Pham,
  Quoc~V. Le, Yun{-}Hsuan Sung, Zhen Li, and Tom Duerig.
\newblock Scaling up visual and vision-language representation learning with
  noisy text supervision.
\newblock In Marina Meila and Tong Zhang, editors, {\em Proceedings of the 38th
  International Conference on Machine Learning, {ICML} 2021, 18-24 July 2021,
  Virtual Event}, volume 139 of {\em Proceedings of Machine Learning Research},
  pages 4904--4916. {PMLR}, 2021.

\bibitem{JiaTCCBHL22}
Menglin Jia, Luming Tang, Bor{-}Chun Chen, Claire Cardie, Serge~J. Belongie,
  Bharath Hariharan, and Ser{-}Nam Lim.
\newblock Visual prompt tuning.
\newblock In Shai Avidan, Gabriel~J. Brostow, Moustapha Ciss{\'{e}},
  Giovanni~Maria Farinella, and Tal Hassner, editors, {\em Computer Vision -
  {ECCV} 2022 - 17th European Conference, Tel Aviv, Israel, October 23-27,
  2022, Proceedings, Part {XXXIII}}, volume 13693 of {\em Lecture Notes in
  Computer Science}, pages 709--727. Springer, 2022.

\bibitem{KimSK21}
Wonjae Kim, Bokyung Son, and Ildoo Kim.
\newblock Vilt: Vision-and-language transformer without convolution or region
  supervision.
\newblock In Marina Meila and Tong Zhang, editors, {\em Proceedings of the 38th
  International Conference on Machine Learning, {ICML} 2021, 18-24 July 2021,
  Virtual Event}, volume 139 of {\em Proceedings of Machine Learning Research},
  pages 5583--5594. {PMLR}, 2021.

\bibitem{Krause0DF13}
Jonathan Krause, Michael Stark, Jia Deng, and Li Fei{-}Fei.
\newblock 3d object representations for fine-grained categorization.
\newblock In {\em 2013 {IEEE} International Conference on Computer Vision
  Workshops, {ICCV} Workshops 2013, Sydney, Australia, December 1-8, 2013},
  pages 554--561. {IEEE} Computer Society, 2013.

\bibitem{abs-2107-13586}
Pengfei Liu, Weizhe Yuan, Jinlan Fu, Zhengbao Jiang, Hiroaki Hayashi, and
  Graham Neubig.
\newblock Pre-train, prompt, and predict: {A} systematic survey of prompting
  methods in natural language processing.
\newblock {\em CoRR}, abs/2107.13586, 2021.

\bibitem{LuBPL19}
Jiasen Lu, Dhruv Batra, Devi Parikh, and Stefan Lee.
\newblock Vilbert: Pretraining task-agnostic visiolinguistic representations
  for vision-and-language tasks.
\newblock In Hanna~M. Wallach, Hugo Larochelle, Alina Beygelzimer, Florence
  d'Alch{\'{e}}{-}Buc, Emily~B. Fox, and Roman Garnett, editors, {\em Advances
  in Neural Information Processing Systems 32: Annual Conference on Neural
  Information Processing Systems 2019, NeurIPS 2019, December 8-14, 2019,
  Vancouver, BC, Canada}, pages 13--23, 2019.

\bibitem{proda}
Yuning Lu, Jianzhuang Liu, Yonggang Zhang, Yajing Liu, and Xinmei Tian.
\newblock Prompt distribution learning.
\newblock In {\em {IEEE/CVF} Conference on Computer Vision and Pattern
  Recognition, {CVPR} 2022, New Orleans, LA, USA, June 18-24, 2022}, pages
  5196--5205. {IEEE}, 2022.

\bibitem{MajiRKBV13}
Subhransu Maji, Esa Rahtu, Juho Kannala, Matthew~B. Blaschko, and Andrea
  Vedaldi.
\newblock Fine-grained visual classification of aircraft.
\newblock {\em CoRR}, abs/1306.5151, 2013.

\bibitem{NilsbackZ08}
Maria{-}Elena Nilsback and Andrew Zisserman.
\newblock Automated flower classification over a large number of classes.
\newblock In {\em Sixth Indian Conference on Computer Vision, Graphics {\&}
  Image Processing, {ICVGIP} 2008, Bhubaneswar, India, 16-19 December 2008},
  pages 722--729. {IEEE} Computer Society, 2008.

\bibitem{ParkhiVZJ12}
Omkar~M. Parkhi, Andrea Vedaldi, Andrew Zisserman, and C.~V. Jawahar.
\newblock Cats and dogs.
\newblock In {\em 2012 {IEEE} Conference on Computer Vision and Pattern
  Recognition, Providence, RI, USA, June 16-21, 2012}, pages 3498--3505. {IEEE}
  Computer Society, 2012.

\bibitem{PetroniRRLBWM19}
Fabio Petroni, Tim Rockt{\"{a}}schel, Sebastian Riedel, Patrick S.~H. Lewis,
  Anton Bakhtin, Yuxiang Wu, and Alexander~H. Miller.
\newblock Language models as knowledge bases?
\newblock In Kentaro Inui, Jing Jiang, Vincent Ng, and Xiaojun Wan, editors,
  {\em Proceedings of the 2019 Conference on Empirical Methods in Natural
  Language Processing and the 9th International Joint Conference on Natural
  Language Processing, {EMNLP-IJCNLP} 2019, Hong Kong, China, November 3-7,
  2019}, pages 2463--2473. Association for Computational Linguistics, 2019.

\bibitem{RadfordKHRGASAM21}
Alec Radford, Jong~Wook Kim, Chris Hallacy, Aditya Ramesh, Gabriel Goh,
  Sandhini Agarwal, Girish Sastry, Amanda Askell, Pamela Mishkin, Jack Clark,
  Gretchen Krueger, and Ilya Sutskever.
\newblock Learning transferable visual models from natural language
  supervision.
\newblock In Marina Meila and Tong Zhang, editors, {\em Proceedings of the 38th
  International Conference on Machine Learning, {ICML} 2021, 18-24 July 2021,
  Virtual Event}, volume 139 of {\em Proceedings of Machine Learning Research},
  pages 8748--8763. {PMLR}, 2021.

\bibitem{RaoZ0TZH0L22}
Yongming Rao, Wenliang Zhao, Guangyi Chen, Yansong Tang, Zheng Zhu, Guan Huang,
  Jie Zhou, and Jiwen Lu.
\newblock Denseclip: Language-guided dense prediction with context-aware
  prompting.
\newblock In {\em {IEEE/CVF} Conference on Computer Vision and Pattern
  Recognition, {CVPR} 2022, New Orleans, LA, USA, June 18-24, 2022}, pages
  18061--18070. {IEEE}, 2022.

\bibitem{RechtRSS19}
Benjamin Recht, Rebecca Roelofs, Ludwig Schmidt, and Vaishaal Shankar.
\newblock Do imagenet classifiers generalize to imagenet?
\newblock In Kamalika Chaudhuri and Ruslan Salakhutdinov, editors, {\em
  Proceedings of the 36th International Conference on Machine Learning, {ICML}
  2019, 9-15 June 2019, Long Beach, California, {USA}}, volume~97 of {\em
  Proceedings of Machine Learning Research}, pages 5389--5400. {PMLR}, 2019.

\bibitem{abs-1212-0402}
Khurram Soomro, Amir~Roshan Zamir, and Mubarak Shah.
\newblock {UCF101:} {A} dataset of 101 human actions classes from videos in the
  wild.
\newblock {\em CoRR}, abs/1212.0402, 2012.

\bibitem{TsimpoukelliMCE21}
Maria Tsimpoukelli, Jacob Menick, Serkan Cabi, S.~M.~Ali Eslami, Oriol Vinyals,
  and Felix Hill.
\newblock Multimodal few-shot learning with frozen language models.
\newblock In Marc'Aurelio Ranzato, Alina Beygelzimer, Yann~N. Dauphin, Percy
  Liang, and Jennifer~Wortman Vaughan, editors, {\em Advances in Neural
  Information Processing Systems 34: Annual Conference on Neural Information
  Processing Systems 2021, NeurIPS 2021, December 6-14, 2021, virtual}, pages
  200--212, 2021.

\bibitem{VaswaniSPUJGKP17}
Ashish Vaswani, Noam Shazeer, Niki Parmar, Jakob Uszkoreit, Llion Jones,
  Aidan~N. Gomez, Lukasz Kaiser, and Illia Polosukhin.
\newblock Attention is all you need.
\newblock In Isabelle Guyon, Ulrike von Luxburg, Samy Bengio, Hanna~M. Wallach,
  Rob Fergus, S.~V.~N. Vishwanathan, and Roman Garnett, editors, {\em Advances
  in Neural Information Processing Systems 30: Annual Conference on Neural
  Information Processing Systems 2017, December 4-9, 2017, Long Beach, CA,
  {USA}}, pages 5998--6008, 2017.

\bibitem{WangGLX19}
Haohan Wang, Songwei Ge, Zachary~C. Lipton, and Eric~P. Xing.
\newblock Learning robust global representations by penalizing local predictive
  power.
\newblock In Hanna~M. Wallach, Hugo Larochelle, Alina Beygelzimer, Florence
  d'Alch{\'{e}}{-}Buc, Emily~B. Fox, and Roman Garnett, editors, {\em Advances
  in Neural Information Processing Systems 32: Annual Conference on Neural
  Information Processing Systems 2019, NeurIPS 2019, December 8-14, 2019,
  Vancouver, BC, Canada}, pages 10506--10518, 2019.

\bibitem{WangYYDT022}
Zirui Wang, Jiahui Yu, Adams~Wei Yu, Zihang Dai, Yulia Tsvetkov, and Yuan Cao.
\newblock Simvlm: Simple visual language model pretraining with weak
  supervision.
\newblock In {\em The Tenth International Conference on Learning
  Representations, {ICLR} 2022, Virtual Event, April 25-29, 2022}.
  OpenReview.net, 2022.

\bibitem{XiaoHEOT10}
Jianxiong Xiao, James Hays, Krista~A. Ehinger, Aude Oliva, and Antonio
  Torralba.
\newblock {SUN} database: Large-scale scene recognition from abbey to zoo.
\newblock In {\em The Twenty-Third {IEEE} Conference on Computer Vision and
  Pattern Recognition, {CVPR} 2010, San Francisco, CA, USA, 13-18 June 2010},
  pages 3485--3492. {IEEE} Computer Society, 2010.

\bibitem{abs-2109-11797}
Yuan Yao, Ao Zhang, Zhengyan Zhang, Zhiyuan Liu, Tat{-}Seng Chua, and Maosong
  Sun.
\newblock {CPT:} colorful prompt tuning for pre-trained vision-language models.
\newblock {\em CoRR}, abs/2109.11797, 2021.

\bibitem{abs-2210-07225}
Yuhang Zang, Wei Li, Kaiyang Zhou, Chen Huang, and Chen~Change Loy.
\newblock Unified vision and language prompt learning.
\newblock {\em CoRR}, abs/2210.07225, 2022.

\bibitem{ZhouYL022}
Kaiyang Zhou, Jingkang Yang, Chen~Change Loy, and Ziwei Liu.
\newblock Conditional prompt learning for vision-language models.
\newblock In {\em {IEEE/CVF} Conference on Computer Vision and Pattern
  Recognition, {CVPR} 2022, New Orleans, LA, USA, June 18-24, 2022}, pages
  16795--16804. {IEEE}, 2022.

\bibitem{ZhouYLL22}
Kaiyang Zhou, Jingkang Yang, Chen~Change Loy, and Ziwei Liu.
\newblock Learning to prompt for vision-language models.
\newblock {\em Int. J. Comput. Vis.}, 130(9):2337--2348, 2022.

\bibitem{abs-2205-14865}
Beier Zhu, Yulei Niu, Yucheng Han, Yue Wu, and Hanwang Zhang.
\newblock Prompt-aligned gradient for prompt tuning.
\newblock {\em CoRR}, abs/2205.14865, 2022.

\end{thebibliography}
}

\clearpage
\begin{appendices}

\setcounter{table}{0}   
\setcounter{figure}{0}
\renewcommand{\thetable}{A\arabic{table}}
\renewcommand{\thefigure}{A\arabic{figure}}
\maketitle

\section{Comparison for Cross-Dataset Transfer}
Similar to CoOp~\cite{ZhouYLL22} and CoCoOp~\cite{ZhouYL022}, we also evaluate the generalizability of the KgCoOp by applying the learnable prompts inferred from the source dataset (ImageNet) on the other downstream dataset.
The related results are shown in Table~\ref{tab:cd}.
As shown in Table~\ref{tab:cd}, CoCoOp obtains the best average performance of all existing methods.
The reason is that the prompts in CoCoOp are a combination of textual prompts and visual descriptions, leading to CoCoOp having high generalizability on unseen datasets.
However, CoCoOp is a time-consuming method.
Different from CoCoOp, CoOp, ProGrad~\cite{abs-2205-14865} and the proposed KgCoOp only use textual-based prompts.
Compared to CoOp and ProGrad, KgCoOp obtains a higher performance on almost all datasets except EuroSAT.
The superior performance proves that the proposed KgCoOp has a high generalizability for cross-dataset transfer.

\section{Effect of Context Length}
\begin{figure}
  \centering
   \includegraphics[width=1.0\linewidth]{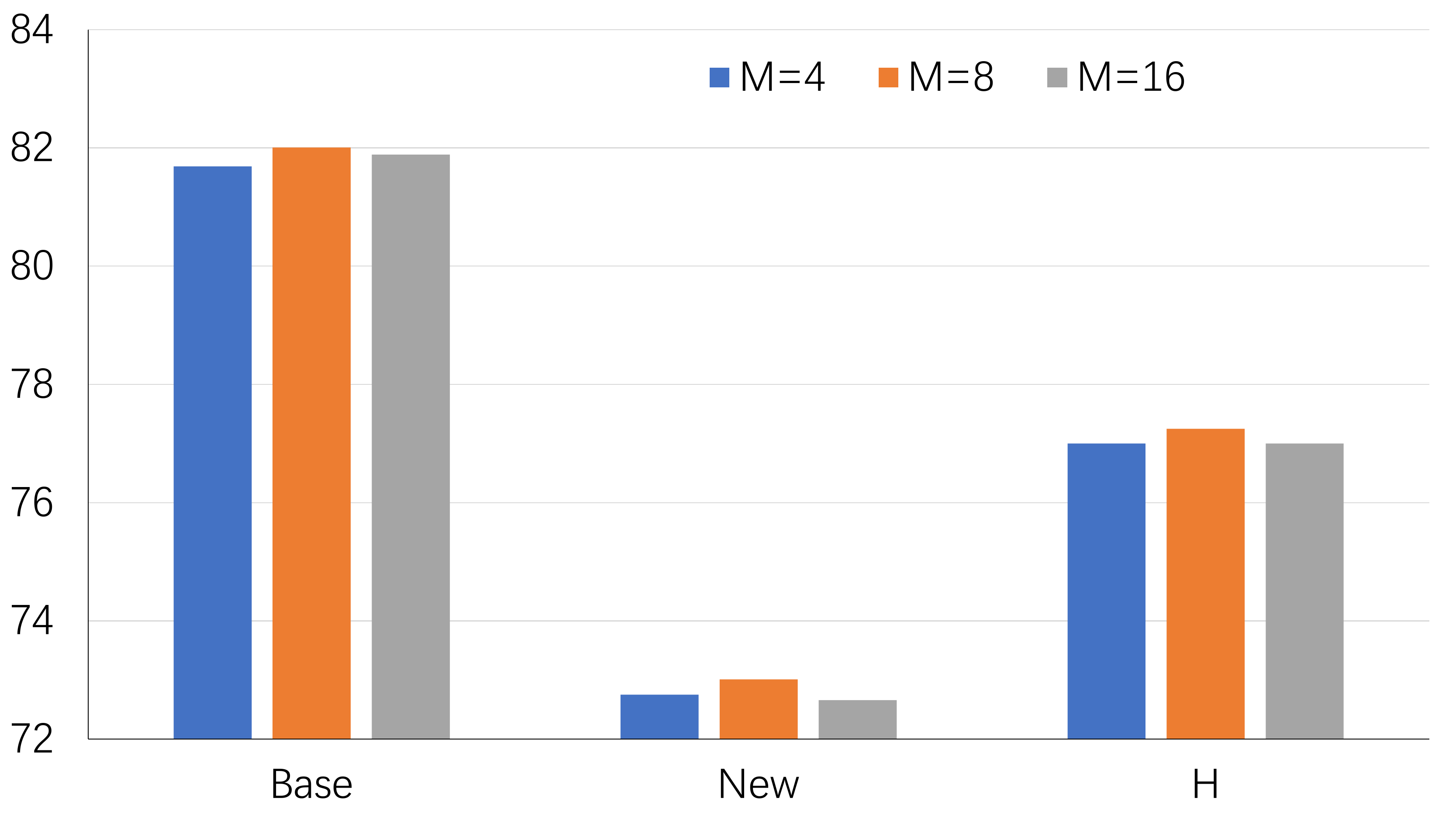}
   \caption{ Effect of context length.}
   \label{fig:cl}
\end{figure}

For the learnable prompts, the context length is a critical aspect.
We thus analyze the effect of the context length in the base-to-new generalization setting with the backbone of ViT-16/B.
Similar to CoOp~\cite{ZhouYLL22}, we study 4, 8, and 16 context tokens.
For the context length of 8 and 16, the prompt is initialized with ``X X ... X a photo of a [Class ]".
The averaging performance on 11 datasets is summarized in Figure~\ref{fig:cl}.
We can observe that setting the context length as 8 obtains a higher performance than the other two settings on all three metric terms.
Furthermore, the learning prompt with lengths of 4 and 16 obtain similar performance.
However, for making a fair comparison with CoOp and CoCoOp, the context length is set as 4 in our final model.

\begin{table}
\caption{Comparison in the cross-dataset transfer learning by learning the prompts from ImageNet(16-shot samples) with ViT-16/B, and evaluating on the other 10 datasets. ``tp" denotes the ``textual prompt", and ``v" denotes the visual information of each instance.}
\label{tab:cd}
\centering
\scriptsize
\begin{tabular}{c|l|l|lll}
\toprule
\multicolumn{1}{l|}{}       & Methods      & \multicolumn{1}{c|}{CoCoOp} & \multicolumn{1}{c}{CoOp} & \multicolumn{1}{c}{ProGrad} & \multicolumn{1}{c}{KgCoOp} \\
\midrule
\multicolumn{1}{l|}{}       & Prompts       & \multicolumn{1}{c|}{tp+v}  & \multicolumn{3}{c}{tp}                                                              \\
\midrule
\multicolumn{1}{l|}{Source} & ImageNet      & 71.02                      & 71.51                    & \textbf{72.24}                       & 70.66                      \\
\midrule
\multirow{11}{*}{Targets}  & Caltech101    & 94.43                      & 93.70                    & 91.52                       & \textbf{93.92}                      \\
                           & OxfordPets    & 90.14                      & 89.14                    & 89.64                       & \textbf{89.83}                      \\
                           & StandfordCars & 65.32                      & 64.51                    & 62.39                       & \textbf{65.41}                      \\
                           & Flowers       & 71.88                      & 68.71                    & 67.87                       & \textbf{70.01}                      \\
                           & Food101       & 86.06                      & 85.30                    & 85.40                       & \textbf{86.36}                      \\
                           & FGVCAircraft  & 22.94                      & 18.47                    & 20.61                       & \textbf{22.51}                      \\
                           & SUN397        & 67.36                      & 64.15                    & 62.47                       & \textbf{66.16}                      \\
                           & DTD           & 45.73                      & 41.92                    & 39.42                       & \textbf{46.35}                      \\
                           & EuroSAT       & 45.37                      & \textbf{46.39}                    & 43.46                       & 46.04                      \\
                           & UCF101        & 68.21                      & 66.55                    & 64.29                       & \textbf{68.50}                      \\
\cline{2-6}
                           & Avg.          & 65.74                      & 63.88                    & 62.71                       & \textbf{65.51}                     \\
\bottomrule
\end{tabular}
\end{table}

\section{Effect of Initialization}
\begin{figure}
  \centering
   \includegraphics[width=1.0\linewidth]{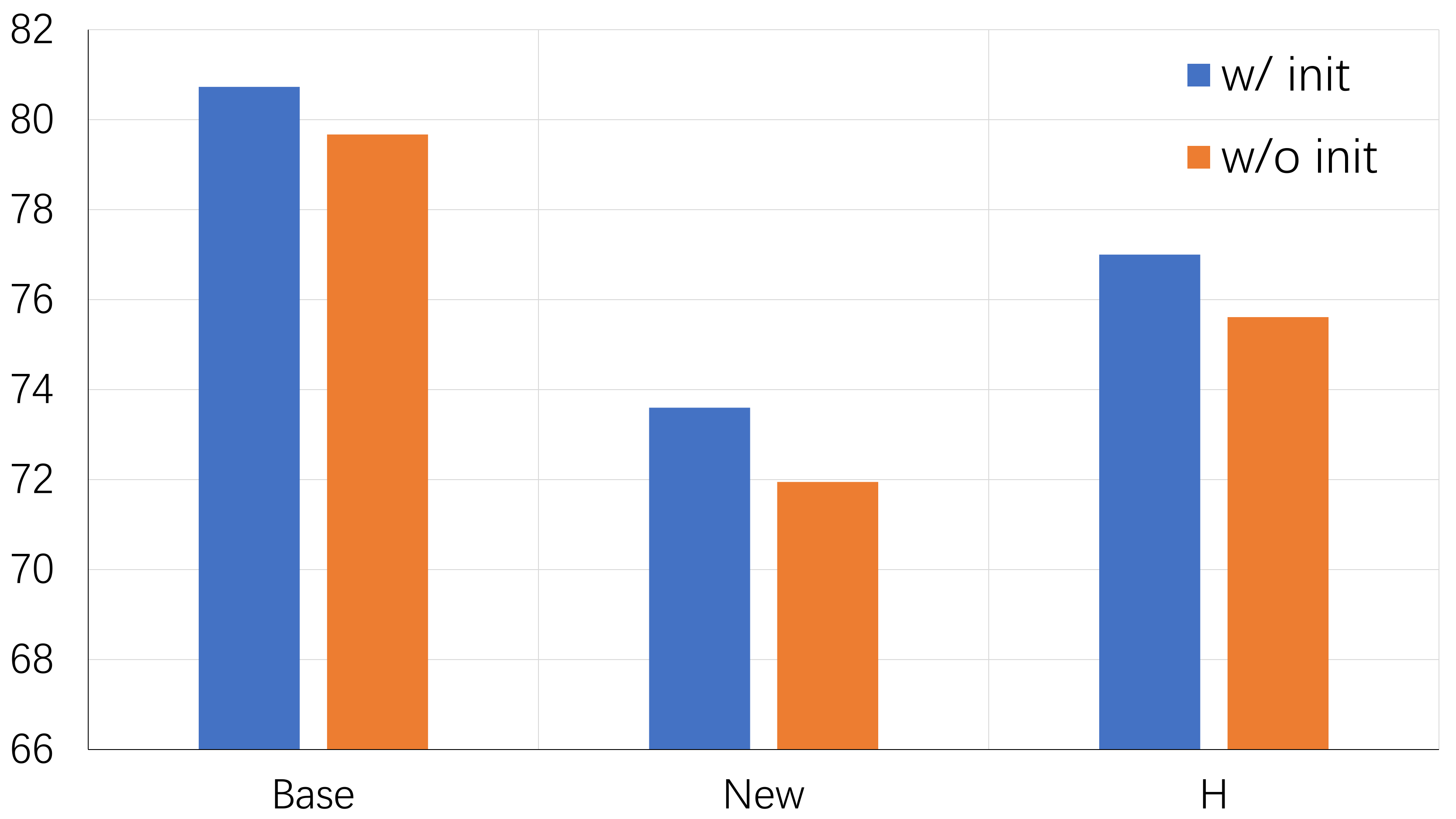}
   \caption{ Effect of initialization.}
   \label{fig:init}
\end{figure}

To verify the impact of initialization for prompt tuning, we conduct a comparison based on the word embeddings-based initialization(`w/ init') and random initialization(`w/o init').
The random initialization applies a zero-mean Gaussian distribution with 0.02 standard deviation to initialize the prompt tokens, and the word embeddings-based initialization uses the ``a photo of a" to initialize the prompt tokens.
The averaging performance on 11 datasets is summarized in Figure~\ref{fig:init}.
We can observe that using the word embedding-based initialization obtains a higher performance in all three terms than random initialization.

\begin{table*}[]
\caption{Comparison in the base-to-new setting with different $K$-shot samples in terms of the average performance among all 11 datasets and backbones(ViT-B/16 and ResNet-50).}
\label{tab:k}
\centering
\small
\begin{tabular}{l|l|ccc||ccc||ccc}
\toprule
  Backbones & Methods& \multicolumn{3}{c||}{$K$=4}                                            & \multicolumn{3}{c||}{$K$=8}                                            & \multicolumn{3}{c}{$K$=16}                                           \\ \cline{3-11} 
        & & Base                 & New                  & H                    & Base                 & New                  & H                    & Base                 & New                  & H                    \\
\midrule
& CoOp    & 78.43                & 68.03                & 72.44                & 80.73                & 68.39                & 73.5                 & 82.63                & 67.99                & 74.60                \\
& CoCoOp  & 76.72                & \textbf{73.34}                & 74.85                & 78.56                & 72.0                 & 74.9                 & 80.47                & 71.69                & 75.83                \\
ViT-B/16& ProGrad & 79.18                & 71.14                & 74.62                & \textbf{80.62}                & 71.02                & 75.2                 & \textbf{82.48}                & 70.75                & 76.16                \\
& KgCoOp     & \textbf{79.92}                & 73.11                & \textbf{75.90}                &      78.36  &    \textbf{73.89}         &     \textbf{76.06}            & 80.73                & \textbf{73.6}                 & \textbf{77.0}                 \\
\midrule
\midrule
& CoOp    & 72.06   & 59.69                & 65.29               & 74.72                & 58.05               & 65.34                & 77.24                & 57.4                & 65.86                \\
& CoCoOp  & 71.39   & 65.74                & 68.45                & 73.4               & 66.42                 & 69.29                &  75.2&          63.64   &       68.9     \\
ResNet-50& ProGrad & \textbf{73.88}               & 64.95                & 69.13                & \textbf{76.25}                & 64.74                & 70.03                 & \textbf{77.98}                & 64.41                & 69.94                \\
& KgCoOp     & 72.42             & \textbf{68.00 }               & \textbf{70.14}                & 74.08 &     \textbf{67.86} &    \textbf{70.84}  &    75.51 &  \textbf{67.53} &  \textbf{71.30}  \\
\bottomrule
\end{tabular}
\end{table*}

\begin{table}
\caption{Effect of hand-crafted prompts. }
\label{tab:hcp}
\centering
\small
\begin{tabular}{c|ccccc}
\toprule
Methods & CoOp & CoCoOp & ProGrad & T1 & T2\\ 
\midrule
H & 74.60& 75.83& 76.16& 76.02& 76.85\\
\midrule
\midrule
Methods & T3 & T4 & T5 &T6 &\\ 
\midrule
H & 76.23& 76.71&76.12&77.0&\\
\bottomrule
\end{tabular}
\end{table}

\section{Effect of hand-crafted prompts}
As different hand-crafted prompts would provide different knowledge to constrain the prompt tuning, we thus evaluate the effect of different hand-crafted prompts.
Evaluation on six hand-crafted prompts shows in Table~\ref{tab:hcp}, \emph{i.e.,} T1:`\{\}'; T2:`a photo of a \{\}'; T3:`itap of a \{\}'; T4:`a photo of the large \{\}'; T5:`a \{\} in a video game'; T6:`a photo of a \{\}, a type of \{\}'.
Although different hand-crafted prompots have achieved different performnce, we observe that T1 without using any prompts obtains the performance of 76.02\%.
Furthermore, the more information given by the hand-crafted prompts, the higher performance, \emph{e.g.,} T6 obtains the highest performance.

\begin{table}
\caption{Comparison of different measurement methods on the average performance of all 11 datasets in the base-to-new setting.}
\label{tab:mm}
\centering
\begin{tabular}{l|ccc}
\toprule
Methods & Base & New & H\\
\midrule
Baseline(CoOp) & 82.63 & 67.99 & 74.60 \\
\midrule
CoOp+$\mathcal{L}_{pt}$ & 78.84 & 70.67 & 74.53 \\
CoOp+$\mathcal{L}_{kl}$ & 80.42 & 72.43 & 76.22 \\
CoOp+$\mathcal{L}_{kg}$ & 80.73 & 73.6  & 77.0\\
\bottomrule
\end{tabular}
\end{table}
\section{How to reduce the discrepancy between special knowledge and general knowledge?}
The key insight of our work is to reduce the discrepancy between special knowledge and general knowledge for improving the generability of unseen datasets.
In KgCoOp, $\mathcal{L}_{kg}$ is used to minimize the distance between the general textual embeddings and specific textual embeddings for reducing the discrepancy.
For the CoOp-based methods, they exist other two ways to measure the discrepancy between special knowledge and general knowledge besides $\mathcal{L}_{kg}$: 1) $\mathcal{L}_{pt}$:reducing the distance between the tokens of the learnable prompts and the fixed prompts; 2) $\mathcal{L}_{kl}$: using the Kullback-Leibler divergence measure the consistency between the predictions generated by the general textual embeddings and specific textual embeddings.
We thus conduct a comparison among all three methods and summarize the results in Table~\ref{tab:mm}.
As shown in Table~\ref{tab:mm}, using $\mathcal{L}_{pt}$ obtains a worse performance of $H$ than CoOp, demonstrating the direct constrain of the similarity between prompts is not a reasonable way.
Different from $\mathcal{L}_{pt}$, $\mathcal{L}_{kg}$ and $\mathcal{L}_{kl}$ both obtain a higher performance than CoOp.
Furthermore, the proposed KgCoOp using $\mathcal{L}_{kg}$ obtains the best performance in the terms of \emph{New} and \emph{H}.
The superior performance proves that it is reasonable to mitigate knowledge forgetting by minimizing the distance between embeddings.

\section{Failure cases}
Similar to ProGrad, we analyze the failure cases where KgCoOp predict incorrectly but CoOp gives right predictions.
Specifically, we count the percentage of the failure cases that zero-shot CLIP models also fails in Figure~\ref{fig:fa}.
We observe that a high proportion of the faiure cases are mis-classified by CoOp model.

\begin{figure}
  \centering
   \includegraphics[width=1.0\linewidth]{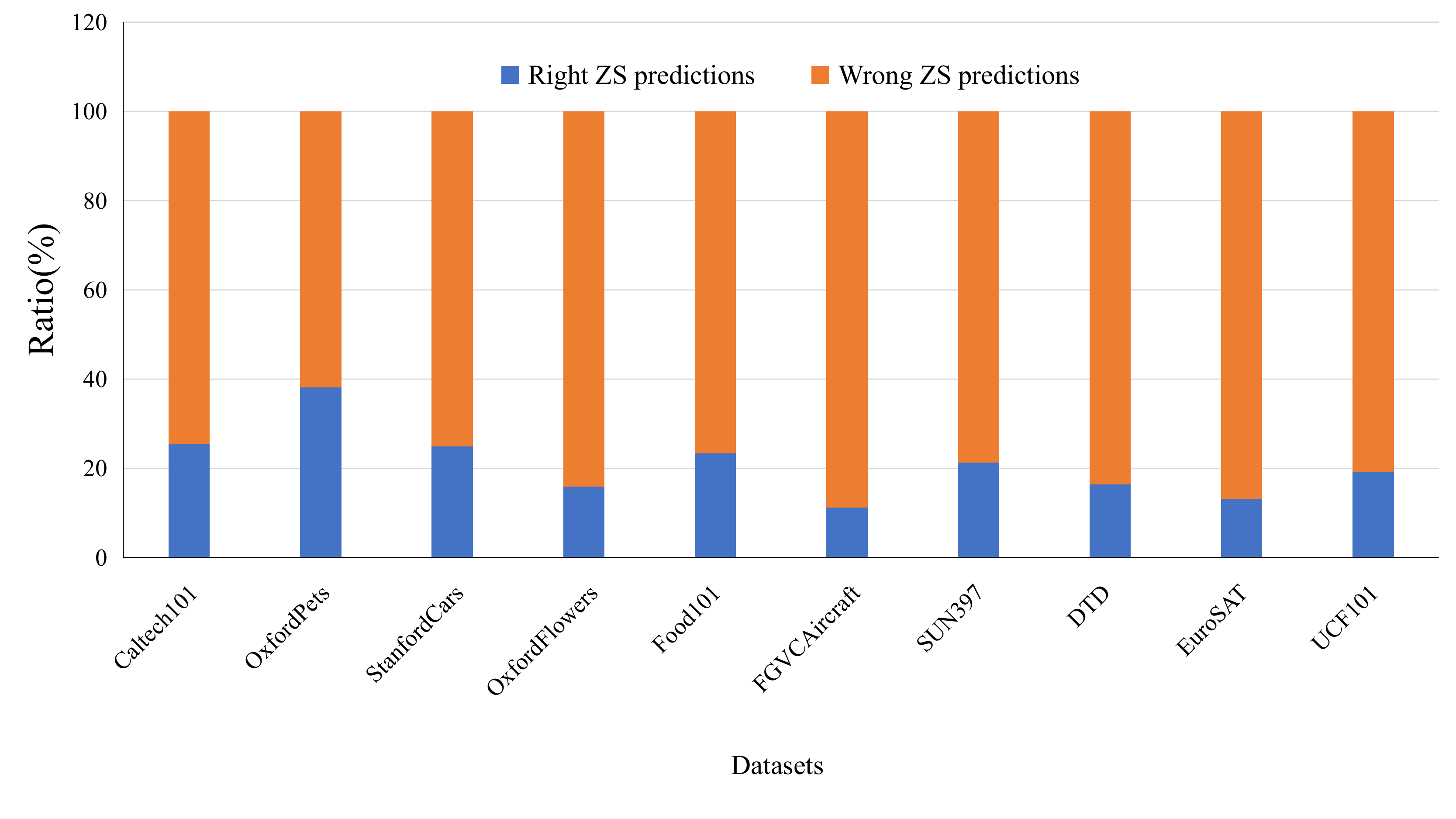}
   \caption{Failure cases analysis. We evaluate the distribution of samples that are mis-classified by KgCoOp but correctly classified by CoOp models}
   \label{fig:fa}
\end{figure}

\section{Disscussion about the generalization on new class}
\begin{figure}
  \centering
   \includegraphics[width=1.0\linewidth]{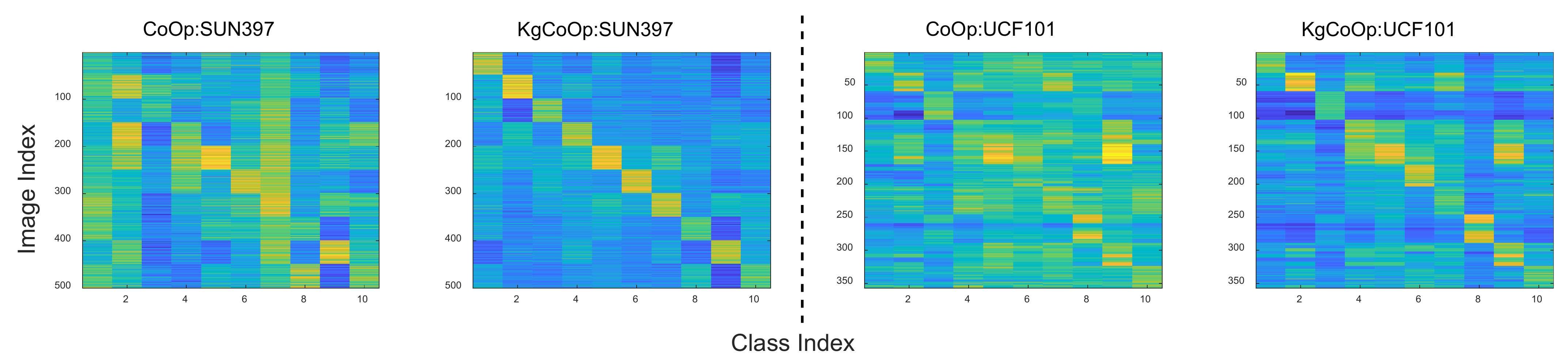}
   \caption{Confusion matrix of the prediction.ls}
   \label{fig:fa}
\label{fig:map}
\end{figure}

As show in Table~\ref{tab:k}, the proposed methods obtains the lower performance on the new class.
The reason is that the domain discrepancy between seen and new classes affects the hardness of generalization to new classes.
Specially, from the Table 3 in the paper, CoOp obtains more than 10\% \emph{New} performance drop on DTD, EuroSAT, and UCF101 datasets.
The reason is that the new classes have a serious domain gap with the seen classes, making the learned prompt biased to the new classes (CoOp in Fig.~\ref{fig:map}).
KgCoOp constrains the learnable prompts to contain the general knowledge in CLIP and discriminative to a new class(Fig.~\ref{fig:map}).
Therefore, KgCoOp significantly improves CoOp for the new classes on those three datasets.

\section{Detailed Results}
To verify the effectiveness of the proposed KgCoOp, we compare KgCoOp with existing CoOp-based methods, \emph{i.e,} CoOp~\cite{ZhouYLL22}, CoCoOp~\cite{ZhouYL022}, and ProGrad~\cite{abs-2205-14865}, based on different backbones and different $K$-shot samples.
Specifily, the CNN-based model ResNet-50~\cite{HeZRS16} and the transformer-based model ViT-B/16~\cite{DosovitskiyB0WZ21} are applied as the visual encoder to extract the image's description.
Furthermore, three types of few-shot settings, \emph{i.e.,} 4-shot, 8-shot, and 16-shot, are conducted for comparison.
The summarized averaged results are shown in Table~\ref{tab:k}.
The detailed results of the backbone of ViT-B/16 are shown in Table~\ref{tab:vit-4} and Table~\ref{tab:vit-8} for 4-shot and 8-shot settings.
For ResNet-50, the results of 4-shot, 8-shot, and 16-shot settings are shown in Table~\ref{tab:resnet_4}, Table~\ref{tab:resnet_8}, and Table~\ref{tab:resnet_8}, respectively.

\begin{table*}
\caption{Comparison with existing methods in the base-to-new generalization based on the  \textbf{ViT-B/16} and \textbf{4-shot} settings. The context length $M$ is 4 for prompot-based methods. H: Harmonic mean.}
\label{tab:vit-4}
\centering
\small
\begin{tabular}{l|lll|lll|lll|lll}
\toprule
              & \multicolumn{3}{c|}{CoOp}                                                   & \multicolumn{3}{c|}{CoCoOp}                                                 & \multicolumn{3}{c|}{ProGrad}                                                & \multicolumn{3}{c}{KgCoOp}                                                 \\\cline{2-13}
Datasets              & \multicolumn{1}{c}{Base} & \multicolumn{1}{c}{New} & \multicolumn{1}{c|}{H} & \multicolumn{1}{c}{Base} & \multicolumn{1}{c}{New} & \multicolumn{1}{c|}{H} & \multicolumn{1}{c}{Base} & \multicolumn{1}{c}{New} & \multicolumn{1}{c|}{H} & \multicolumn{1}{c}{Base} & \multicolumn{1}{c}{New} & \multicolumn{1}{c}{H} \\
\midrule
ImageNet      & 73.60                    & 63.29                   & 68.06                 & 75.46                    & 69.58                   & 72.40                 & 74.24                    & 65.47                   & 69.58                 & 74.87                    & 69.09                   & 71.86                 \\
Caltech101    & 97.27                    & 93.01                   & 95.09                 & 97.25                    & 94.90                   & 96.06                 & 97.37                    & 93.92                   & 95.61                 & 97.53                    & 94.43                   & 95.95                 \\
OxfordPets    & 93.33                    & 95.69                   & 94.50                 & 94.59                    & 96.75                   & 95.66                 & 94.08                    & 97.63                   & 95.82                 & 94.68                    & 97.58                   & 96.11                 \\
StandfordCars & 70.92                    & 69.38                   & 70.14                 & 67.71                    & 75.37                   & 71.33                 & 72.69                    & 69.88                   & 71.26                 & 69.25                    & 74.98                   & 72.00                 \\
Flowers       & 92.50                    & 70.12                   & 79.77                 & 84.75                    & 73.85                   & 78.93                 & 92.46                    & 72.69                   & 81.39                 & 91.30                    & 75.34                   & 82.56                 \\
Food101       & 86.79                    & 89.06                   & 87.91                 & 89.79                    & 90.99                   & 90.39                 & 88.91                    & 90.18                   & 89.54                 & 90.30                    & 91.39                   & 90.84                 \\
FGVCAircraft  & 33.21                    & 28.57                   & 30.72                 & 32.07                    & 33.93                   & 32.97                 & 33.73                    & 30.09                   & 31.81                 & 34.21                    & 32.81                   & 33.50                 \\
SUN397        & 76.49                    & 64.56                   & 70.02                 & 77.57                    & 76.96                   & 77.26                 & 77.72                    & 71.93                   & 74.71                 & 78.87                    & 75.64                   & 77.22                 \\
DTD           & 71.26                    & 50.93                   & 59.40                 & 67.44                    & 56.00                   & 61.19                 & 71.06                    & 52.58                   & 60.44                 & 73.65                    & 57.21                   & 64.40                 \\
EuroSAT       & 82.56                    & 53.04                   & 64.59                 & 79.27                    & 65.44                   & 71.69                 & 82.48                    & 56.43                   & 67.01                 & 82.63                    & 59.98                   & 69.51                 \\
UCF101        & 79.97                    & 65.98                   & 72.30                 & 78.01                    & 73.07                   & 75.46                 & 81.30                    & 76.02                   & 78.57                 & 80.80                    & 75.77                   & 78.20                 \\
\midrule
 Avg.             & 78.43                    & 68.03                   & 72.44                 & 76.72                    & 73.35                   & 74.85                 & 79.18                    & 71.14                   & 74.62                 & 78.92                    & 73.11                   & 75.90 \\
\bottomrule               
\end{tabular}
\end{table*}

\begin{table*}
\caption{Comparison with existing methods in the base-to-new generalization based on the \textbf{ViT-B/16} and \textbf{8-shot} settings. The context length $M$ is 4 for prompot-based methods. H: Harmonic mean.}
\label{tab:vit-8}
\centering
\small
\begin{tabular}{l|lll|lll|lll|lll}
\toprule
              & \multicolumn{3}{c|}{CoOp}                                                   & \multicolumn{3}{c|}{CoCoOp}                                                 & \multicolumn{3}{c|}{ProGrad}                                                & \multicolumn{3}{c}{KgCoOp}                                                 \\\cline{2-13}
Datasets              & \multicolumn{1}{c}{Base} & \multicolumn{1}{c}{New} & \multicolumn{1}{c|}{H} & \multicolumn{1}{c}{Base} & \multicolumn{1}{c}{New} & \multicolumn{1}{c|}{H} & \multicolumn{1}{c}{Base} & \multicolumn{1}{c}{New} & \multicolumn{1}{c|}{H} & \multicolumn{1}{c}{Base} & \multicolumn{1}{c}{New} & \multicolumn{1}{c}{H} \\
\midrule
ImageNet      & 75.22                    & 65.91                   & 70.26                 & 75.52                    & 70.28                   & 72.81                 & 75.72                    & 66.76                   & 70.96                 & 75.84                    & 69.33                   & 72.44                 \\
Caltech101    & 97.81                    & 92.58                   & 95.12                 & 97.76                    & 93.63                   & 95.65                 & 98.00                    & 93.38                   & 95.63                 & 97.68                    & 94.10                   & 95.86                 \\
OxfordPets    & 94.19                    & 96.11                   & 95.14                 & 95.50                    & 97.69                   & 96.58                 & 94.47                    & 97.03                   & 95.73                 & 94.81                    & 97.58                   & 96.18                 \\
StandfordCars & 73.20                    & 67.44                   & 70.20                 & 69.70                    & 74.13                   & 71.85                 & 75.08                    & 70.63                   & 72.79                 & 69.66                    & 75.40                   & 72.42                 \\
Flowers       & 96.17                    & 69.41                   & 80.63                 & 92.24                    & 72.77                   & 81.36                 & 93.80                    & 72.20                   & 81.59                 & 87.72                    & 74.75                   & 80.72                 \\
Food101       & 87.27                    & 86.96                   & 87.11                 & 89.60                    & 90.79                   & 90.19                 & 89.48                    & 89.90                   & 89.69                 & 90.46                    & 91.63                   & 91.04                 \\
FGVCAircraft  & 37.01                    & 38.45                   & 37.72                 & 33.71                    & 32.15                   & 32.91                 & 36.89                    & 31.67                   & 34.08                 & 34.53                    & 34.95                   & 34.74                 \\
SUN397        & 78.61                    & 66.25                   & 71.90                 & 78.05                    & 76.29                   & 77.16                 & 79.21                    & 70.77                   & 74.75                 & 79.37                    & 76.85                   & 78.09                 \\
DTD           & 76.97                    & 51.81                   & 61.93                 & 73.03                    & 57.24                   & 64.18                 & 74.42                    & 52.38                   & 61.48                 & 69.72                    & 56.44                   & 62.38                 \\
EuroSAT       & 83.27                    & 50.59                   & 62.94                 & 78.68                    & 56.03                   & 65.45                 & 82.27                    & 58.52                   & 68.39                 & 81.07                    & 63.13                   & 70.98                 \\
UCF101        & 82.85                    & 64.32                   & 72.42                 & 80.40                    & 71.68                   & 75.79                 & 82.61                    & 73.75                   & 77.93                 & 81.16                    & 78.65                   & 79.89                 \\
\midrule
Avg.              & 80.74                    & 68.39                   & 73.51                 & 78.56                    & 72.06                   & 74.90                 & 80.62                    & 71.02                   & 75.21                 & 78.37                    & 73.89                   & 76.06 \\
\bottomrule               
\end{tabular}
\end{table*}

\begin{table*}
\caption{Comparison with existing methods in the base-to-new generalization based on the \textbf{ResNet-50} and \textbf{4-shot} settings. The context length $M$ is 4 for prompot-based methods. H: Harmonic mean.}
\label{tab:resnet_4}
\centering
\small
\begin{tabular}{l|lll|lll|lll|lll}
\toprule
              & \multicolumn{3}{c|}{CoOp}                                                   & \multicolumn{3}{c|}{CoCoOp}                                                 & \multicolumn{3}{c|}{ProGrad}                                                & \multicolumn{3}{c}{KgCoOp}                                                 \\\cline{2-13}
Datasets              & \multicolumn{1}{c}{Base} & \multicolumn{1}{c}{New} & \multicolumn{1}{c|}{H} & \multicolumn{1}{c}{Base} & \multicolumn{1}{c}{New} & \multicolumn{1}{c|}{H} & \multicolumn{1}{c}{Base} & \multicolumn{1}{c}{New} & \multicolumn{1}{c|}{H} & \multicolumn{1}{c}{Base} & \multicolumn{1}{c}{New} & \multicolumn{1}{c}{H} \\
\midrule
ImageNet      & 64.53                    & 54.47                   & 59.07                 & 67.80                    & 62.45                   & 65.02                 & 65.23                    & 55.96                   & 60.24                 & 67.13                    & 61.96                   & 64.44                 \\
Caltech101    & 94.06                    & 87.01                   & 90.40                 & 95.03                    & 90.47                   & 92.69                 & 94.47                    & 89.26                   & 91.79                 & 94.43                    & 91.56                   & 92.97                 \\
OxfordPets    & 87.36                    & 93.49                   & 90.32                 & 91.62                    & 94.99                   & 93.27                 & 91.25                    & 94.93                   & 93.05                 & 92.29                    & 94.13                   & 93.20                 \\
StandfordCars & 61.84                    & 57.25                   & 59.46                 & 60.58                    & 64.78                   & 62.61                 & 64.98                    & 61.92                   & 63.41                 & 60.53                    & 67.42                   & 63.79                 \\
Flowers       & 89.71                    & 57.68                   & 70.21                 & 81.86                    & 71.44                   & 76.30                 & 90.12                    & 68.82                   & 78.04                 & 78.12                    & 72.77                   & 75.35                 \\
Food101       & 77.20                    & 76.85                   & 77.02                 & 83.19                    & 84.53                   & 83.85                 & 81.48                    & 82.54                   & 82.01                 & 83.56                    & 84.86                   & 84.20                 \\
FGVCAircraft  & 22.19                    & 18.36                   & 20.09                 & 22.55                    & 25.03                   & 23.73                 & 23.47                    & 18.44                   & 20.65                 & 22.53                    & 26.83                   & 24.49                 \\
SUN397        & 70.68                    & 60.87                   & 65.41                 & 72.03                    & 71.76                   & 71.89                 & 73.53                    & 67.04                   & 70.14                 & 73.68                    & 71.92                   & 72.79                 \\
DTD           & 64.74                    & 47.18                   & 54.58                 & 61.77                    & 53.34                   & 57.25                 & 67.90                    & 52.94                   & 59.49                 & 66.24                    & 53.54                   & 59.22                 \\
EuroSAT       & 86.39                    & 46.91                   & 60.80                 & 75.60                    & 37.68                   & 50.29                 & 84.74                    & 60.46                   & 70.57                 & 84.87                    & 52.55                   & 64.91                 \\
UCF101        & 73.96                    & 56.53                   & 64.08                 & 73.27                    & 66.70                   & 69.83                 & 75.56                    & 62.13                   & 68.19                 & 73.20                    & 70.43                   & 71.79                 \\
\midrule
  Avg.            & 72.06                    & 59.69                   & 65.29                 & 71.39                    & 65.74                   & 68.45                 & 73.88                    & 64.95                   & 69.13                 & 72.42                    & 68.00                   & 70.14    \\
\bottomrule         
\end{tabular}
\end{table*}

\begin{table*}
\caption{Comparison with existing methods in the base-to-new generalization based on the \textbf{ResNet-50} and \textbf{8-shot} settings. The context length $M$ is 4 for prompot-based methods. H: Harmonic mean.}
\label{tab:resnet_8}
\centering
\small
\begin{tabular}{l|lll|lll|lll|lll}
\toprule
              & \multicolumn{3}{c|}{CoOp}                                                   & \multicolumn{3}{c|}{CoCoOp}                                                 & \multicolumn{3}{c|}{ProGrad}                                                & \multicolumn{3}{c}{KgCoOp}                                                 \\\cline{2-13}
Datasets              & \multicolumn{1}{c}{Base} & \multicolumn{1}{c}{New} & \multicolumn{1}{c|}{H} & \multicolumn{1}{c}{Base} & \multicolumn{1}{c}{New} & \multicolumn{1}{c|}{H} & \multicolumn{1}{c}{Base} & \multicolumn{1}{c}{New} & \multicolumn{1}{c|}{H} & \multicolumn{1}{c}{Base} & \multicolumn{1}{c}{New} & \multicolumn{1}{c}{H} \\
\midrule
ImageNet      & 66.69                    & 57.36                   & 61.67                 & 68.06                    & 62.71                   & 65.28                 & 67.25                    & 57.83                   & 62.19                 & 67.62                    & 62.27                   & 64.83                 \\
Caltech101    & 94.40                    & 83.88                   & 88.83                 & 95.31                    & 91.05                   & 93.13                 & 95.12                    & 88.97                   & 91.94                 & 94.92                    & 91.88                   & 93.38                 \\
OxfordPets    & 90.02                    & 93.36                   & 91.66                 & 92.45                    & 95.73                   & 94.06                 & 91.90                    & 94.59                   & 93.23                 & 92.36                    & 94.37                   & 93.35                 \\
StandfordCars & 65.49                    & 55.89                   & 60.31                 & 61.61                    & 65.98                   & 63.72                 & 68.33                    & 60.10                   & 63.95                 & 60.91                    & 66.55                   & 63.61                 \\
Flowers       & 93.07                    & 57.59                   & 71.15                 & 85.25                    & 68.56                   & 76.00                 & 92.46                    & 67.59                   & 78.09                 & 87.18                    & 72.67                   & 79.27                 \\
Food101       & 78.55                    & 78.03                   & 78.29                 & 84.09                    & 85.37                   & 84.73                 & 82.50                    & 83.36                   & 82.93                 & 83.74                    & 85.21                   & 84.47                 \\
FGVCAircraft  & 25.01                    & 18.04                   & 20.96                 & 23.17                    & 23.60                   & 23.38                 & 27.71                    & 20.58                   & 23.62                 & 24.15                    & 26.83                   & 25.42                 \\
SUN397        & 73.58                    & 60.95                   & 66.67                 & 73.53                    & 72.52                   & 73.02                 & 75.13                    & 67.03                   & 70.85                 & 74.63                    & 72.21                   & 73.40                 \\
DTD           & 71.53                    & 40.34                   & 51.59                 & 68.29                    & 49.76                   & 57.57                 & 71.61                    & 47.58                   & 57.17                 & 69.25                    & 51.57                   & 59.12                 \\
EuroSAT       & 85.88                    & 42.46                   & 56.83                 & 80.43                    & 48.75                   & 60.71                 & 87.45                    & 59.75                   & 70.99                 & 83.87                    & 52.80                   & 64.80                 \\
UCF101        & 77.69                    & 50.64                   & 61.31                 & 75.23                    & 66.54                   & 70.62                 & 79.30                    & 64.81                   & 71.33                 & 76.28                    & 70.18                   & 73.10                 \\
\midrule
 Avg.             & 74.72                    & 58.05                   & 65.34                 & 73.40                    & 66.42                   & 69.29                 & 76.25                    & 64.74                   & 70.03                 & 74.08                    & 67.87                   & 70.84 \\
\bottomrule               
\end{tabular}
\end{table*}

\begin{table*}
\caption{Comparison with existing methods in the base-to-new generalization based on the \textbf{ResNet-50} and \textbf{16-shot} settings. The context length $M$ is 4 for prompot-based methods. H: Harmonic mean.}
\label{tab:resnet_16}
\centering
\small
\begin{tabular}{l|lll|lll|lll|lll}
\toprule
              & \multicolumn{3}{c|}{CoOp}                                                   & \multicolumn{3}{c|}{CoCoOp}                                                 & \multicolumn{3}{c|}{ProGrad}                                                & \multicolumn{3}{c}{KgCoOp}                                                 \\\cline{2-13}
Datasets              & \multicolumn{1}{c}{Base} & \multicolumn{1}{c}{New} & \multicolumn{1}{c|}{H} & \multicolumn{1}{c}{Base} & \multicolumn{1}{c}{New} & \multicolumn{1}{c|}{H} & \multicolumn{1}{c}{Base} & \multicolumn{1}{c}{New} & \multicolumn{1}{c|}{H} & \multicolumn{1}{c}{Base} & \multicolumn{1}{c}{New} & \multicolumn{1}{c}{H} \\
\midrule
ImageNet      & 68.57                    & 58.76                   & 63.29                 & 68.21                    & 62.28                   & 65.11                 & 69.13                    & 57.39                   & 62.72                 & 67.67                    & 62.45                   & 64.96                 \\
Caltech101    & 95.20                    & 87.55                   & 91.21                 & 95.40                    & 90.28                   & 92.77                 & 95.72                    & 89.92                   & 92.73                 & 95.35                    & 91.92                   & 93.60                 \\
OxfordPets    & 90.15                    & 90.70                   & 90.42                 & 92.10                    & 95.81                   & 93.92                 & 92.36                    & 94.48                   & 93.41                 & 92.57                    & 94.61                   & 93.58                 \\
StandfordCars & 68.89                    & 57.13                   & 62.46                 & 63.53                    & 64.46                   & 63.99                 & 71.79                    & 59.36                   & 64.99                 & 63.28                    & 66.92                   & 65.05                 \\
Flowers       & 95.22                    & 59.53                   & 73.26                 & 90.66                    & 67.19                   & 77.18                 & 94.71                    & 68.86                   & 79.74                 & 91.45                    & 71.75                   & 80.41                 \\
Food101       & 81.70                    & 78.13                   & 79.88                 & 84.44                    & 85.80                   & 85.11                 & 83.77                    & 83.74                   & 83.75                 & 83.90                    & 85.23                   & 84.56                 \\
FGVCAircraft  & 28.39                    & 20.02                   & 23.48                 & 23.98                    & 21.05                   & 22.42                 & 30.17                    & 19.70                   & 23.84                 & 24.91                    & 25.69                   & 25.29                 \\
SUN397        & 76.33                    & 62.89                   & 68.96                 & 74.64                    & 72.78                   & 73.70                 & 76.90                    & 68.09                   & 72.23                 & 75.33                    & 72.25                   & 73.76                 \\
DTD           & 75.12                    & 37.08                   & 49.65                 & 71.18                    & 47.42                   & 56.92                 & 73.80                    & 46.38                   & 56.96                 & 74.73                    & 48.39                   & 58.74                 \\
EuroSAT       & 90.25                    & 31.30                   & 46.48                 & 86.13                    & 31.65                   & 46.29                 & 88.44                    & 49.49                   & 63.47                 & 84.28                    & 53.53                   & 65.47                 \\
UCF101        & 79.78                    & 48.31                   & 60.18                 & 76.92                    & 61.38                   & 68.28                 & 81.04                    & 60.07                   & 69.00                 & 77.16                    & 70.13                   & 73.48                 \\
\midrule
Avg.              & 77.24                    & 57.40                   & 65.86                 & 75.20                    & 63.65                   & 68.94                 & 77.98                    & 63.41                   & 69.94                 & 75.51                    & 67.53                   & 71.30\\
\bottomrule                
\end{tabular}
\end{table*}

\end{appendices}

\end{document}